\PassOptionsToPackage{table}{xcolor}
\documentclass{article}

\usepackage{iclr2026_conference,times}
\usepackage[letterpaper,left=0.9in,right=0.9in,top=0.68in,bottom=0.72in]{geometry}
\usepackage{titlesec}

\usepackage[utf8]{inputenc}
\usepackage[T1]{fontenc}
\usepackage{amsmath}
\usepackage{amssymb}
\usepackage{booktabs}
\usepackage{enumitem}
\usepackage{graphicx}
\usepackage{hyperref}
\usepackage{url}
\usepackage{algorithm}
\usepackage{algorithmic}
\usepackage[capitalize,nameinlink]{cleveref}
\usepackage{multirow}
\usepackage{xcolor}
\usepackage{tikz}
\usepackage{tabularx}
\newcolumntype{Y}{>{\centering\arraybackslash}X}

\makeatletter
\DeclareRobustCommand\onedot{\futurelet\@let@token\@onedot}
\def\@onedot{\ifx\@let@token.\else.\null\fi\xspace}
\def\eg{\emph{e.g}\onedot} 
\def\ie{\emph{i.e}\onedot}

\makeatother

\definecolor{brandcolor}{HTML}{174B3B}
\definecolor{frontgray}{HTML}{F1F2F3}
\definecolor{frontdark}{HTML}{4B4B4B}
\usepackage{xspace}
\usepackage{tcolorbox}
\usepackage{caption}
\usepackage{placeins}
\graphicspath{{fig/}}

\hypersetup{
    colorlinks=true,
    linkcolor=blue,
    urlcolor=blue,
    citecolor=blue
}

\newcommand{\methodname}{\texorpdfstring{\textcolor{brandcolor}{LiveVAE}}{LiveVAE}\xspace}

\newtcolorbox{frontmatterbox}{
    colback=frontgray,
    colframe=frontgray,
    boxrule=0pt,
    arc=8pt,
    left=20pt,
    right=20pt,
    top=18pt,
    bottom=18pt,
    width=\textwidth,
    before skip=0pt,
    after skip=0pt
}

\titleformat{\section}{\sffamily\Large\bfseries}{\thesection}{1em}{}
\titleformat{\subsection}{\sffamily\large\bfseries}{\thesubsection}{0.8em}{}
\titleformat{\subsubsection}{\sffamily\normalsize\bfseries}{\thesubsubsection}{0.8em}{}
\titlespacing*{\section}{0pt}{2.0ex plus 0.4ex minus 0.2ex}{1.1ex}
\titlespacing*{\subsection}{0pt}{1.7ex plus 0.3ex minus 0.2ex}{0.8ex}
\titlespacing*{\subsubsection}{0pt}{1.4ex plus 0.3ex minus 0.2ex}{0.6ex}

\newcommand{\papersubtitle}{Causal Window Attention Makes Efficient Streaming Video VAE}
\newcommand{\paperfrontsubtitle}{Causal Window Attention Makes Efficient Streaming Video VAE}
\newcommand{\projectpageurl}{https://livevae.github.io/}
\newcommand{\projectpagelabel}{livevae.github.io}
\newcommand{\codeurl}{https://github.com/MeiGen-AI/LiveVAE}
\newcommand{\codelabel}{MeiGen-AI/LiveVAE}

\title{\methodname: \papersubtitle}

\author{
\textbf{\mbox{Jiachen~An}\textsuperscript{1,2,4,*},
\mbox{Sijie~Zhao}\textsuperscript{2,*},
\mbox{Le~Li}\textsuperscript{2,*},
\mbox{Hongyu~Liu}\textsuperscript{2,3},
\mbox{Zhiyuan~Li}\textsuperscript{1,2,4},
\mbox{Tianyu~Yang}\textsuperscript{2},
\mbox{Xiaoyu~Kong}\textsuperscript{2},
\mbox{Xunliang~Cai}\textsuperscript{2},
\mbox{Yong~Zhang}\textsuperscript{2,$\ddagger$},
\mbox{Chi-Man~Pun}\textsuperscript{4,$\dagger$},
\mbox{Xiaoming~Wei}\textsuperscript{2,$\dagger$},
\mbox{Xiaodong~Cun}\textsuperscript{1,$\dagger$}} \\
\textsuperscript{1}GVC Lab, Great Bay University \quad
\textsuperscript{2}Meituan \quad
\textsuperscript{3}The Hong Kong University of Science and Technology \\
\textsuperscript{4}University of Macau \\[2pt]
{\small \textsuperscript{*}Equal contribution \quad \textsuperscript{$\dagger$}Corresponding authors \quad \textsuperscript{$\ddagger$}Project lead}
}

\newcommand{\frontauthorblock}{%
\raggedright
\textbf{\mbox{Jiachen~An}\textsuperscript{1,2,4,*},
\mbox{Sijie~Zhao}\textsuperscript{2,*},
\mbox{Le~Li}\textsuperscript{2,*},
\mbox{Hongyu~Liu}\textsuperscript{2,3},
\mbox{Zhiyuan~Li}\textsuperscript{1,2,4},
\mbox{Tianyu~Yang}\textsuperscript{2},
\mbox{Xiaoyu~Kong}\textsuperscript{2},
\mbox{Xunliang~Cai}\textsuperscript{2},
\mbox{Yong~Zhang}\textsuperscript{2,$\ddagger$},
\mbox{Chi-Man~Pun}\textsuperscript{4,$\dagger$},
\mbox{Xiaoming~Wei}\textsuperscript{2,$\dagger$},
\mbox{Xiaodong~Cun}\textsuperscript{1,$\dagger$}}\\[4pt]
\textsuperscript{1}GVC Lab, Great Bay University \quad
\textsuperscript{2}Meituan \quad
\textsuperscript{3}The Hong Kong University of Science and Technology \\
\textsuperscript{4}University of Macau \\[2pt]
{\small \textsuperscript{*}Equal contribution \quad \textsuperscript{$\dagger$}Corresponding authors \quad \textsuperscript{$\ddagger$}Project lead}%
}

\newcommand{\frontaffiliationlogos}{%
\includegraphics[height=0.20in]{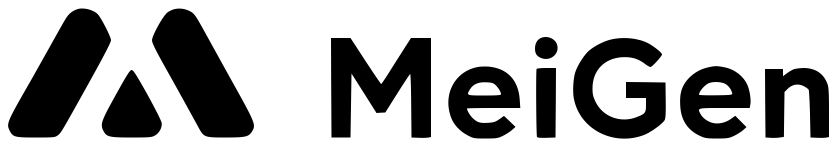}\hspace{0.16in}%
\includegraphics[height=0.20in]{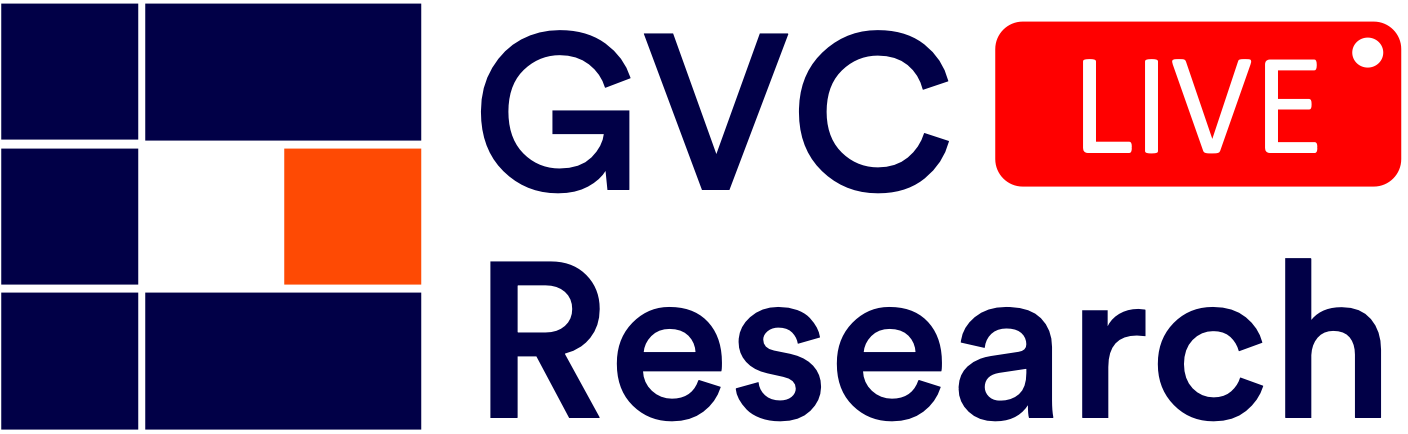}\hspace{0.16in}%
\raisebox{-0.03in}{\includegraphics[height=0.28in]{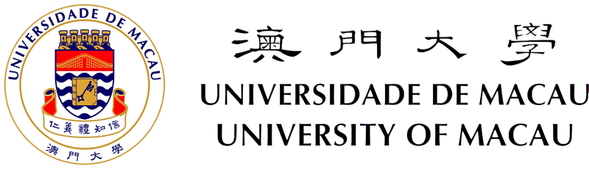}}%
}

\newcommand{\makefrontmatter}{%
\vspace*{0.16in}
\begin{frontmatterbox}
{\sffamily\bfseries\fontsize{28}{31}\selectfont \textcolor{brandcolor}{LiveVAE}\par}
\vspace{0.08in}
{\sffamily\bfseries\Large\color{frontdark}\paperfrontsubtitle\par}
\vspace{0.22in}
{\normalsize\frontauthorblock\par}
\vspace{0.26in}
{\normalsize\paperabstract\par}
\vspace{0.22in}
\noindent
\begin{minipage}[b]{\linewidth}
{\small
\textbf{Project page:} \href{\projectpageurl}{\texttt{\projectpagelabel}}\\
\textbf{Code:} \href{\codeurl}{\texttt{\codelabel}}%
}
\end{minipage}\par\vspace{0.10in}
\noindent
\begin{minipage}[b]{\linewidth}
\raggedleft
\frontaffiliationlogos
\end{minipage}
\end{frontmatterbox}
\vspace{0.22in}
}

\arxivcopy
\renewcommand{\headrulewidth}{0pt}
\renewcommand{\footrulewidth}{0pt}
\AddToShipoutPicture{%
  \AtPageLowerLeft{%
    \raisebox{0.34in}{\makebox[\paperwidth][c]{\thepage}}%
  }%
}

\newcommand{\paperabstract}{%
While video diffusion models have grown faster and increasingly streamable, the video VAE remains a primary latency bottleneck across interactive generation and editing pipelines. Existing VAEs and their acceleration methods rely mainly on heavy convolutions or full frame attention, and most focus only on decoding, leaving encoding costly for streaming conditional generation and editing. This raises a natural question: \textit{Do video VAE encoding and decoding actually require such global spatiotemporal modeling?} By analyzing the effective receptive fields of pretrained video VAEs, we find that \textit{both encoding and decoding a patch rely almost entirely on its spatial and temporal neighbors}. To exploit this strong locality, we introduce LiveVAE, a streaming encoder and decoder built on causal window attention, which cuts per frame compute while keeping the temporal KV cache bounded for streaming. Decoupled encoder/decoder distillation followed by joint finetuning preserves the pretrained latent space, making LiveVAE a drop-in replacement for the original VAE. On MiniMax-H3, LiveVAE achieves $5.2$--$6.2\times$ faster encoding and $3.0$--$3.3\times$ faster decoding than the official VAE with higher PSNR. Paired with our dedicated inference infrastructure, LiveVAE pushes the speedup to $13.4\times$ with almost no quality loss. Code and models will be released.
}

\begin{document}
\makefrontmatter
\section{Introduction}
\label{sec:intro}

\begin{figure}[t]
\centering
\includegraphics[width=\linewidth]{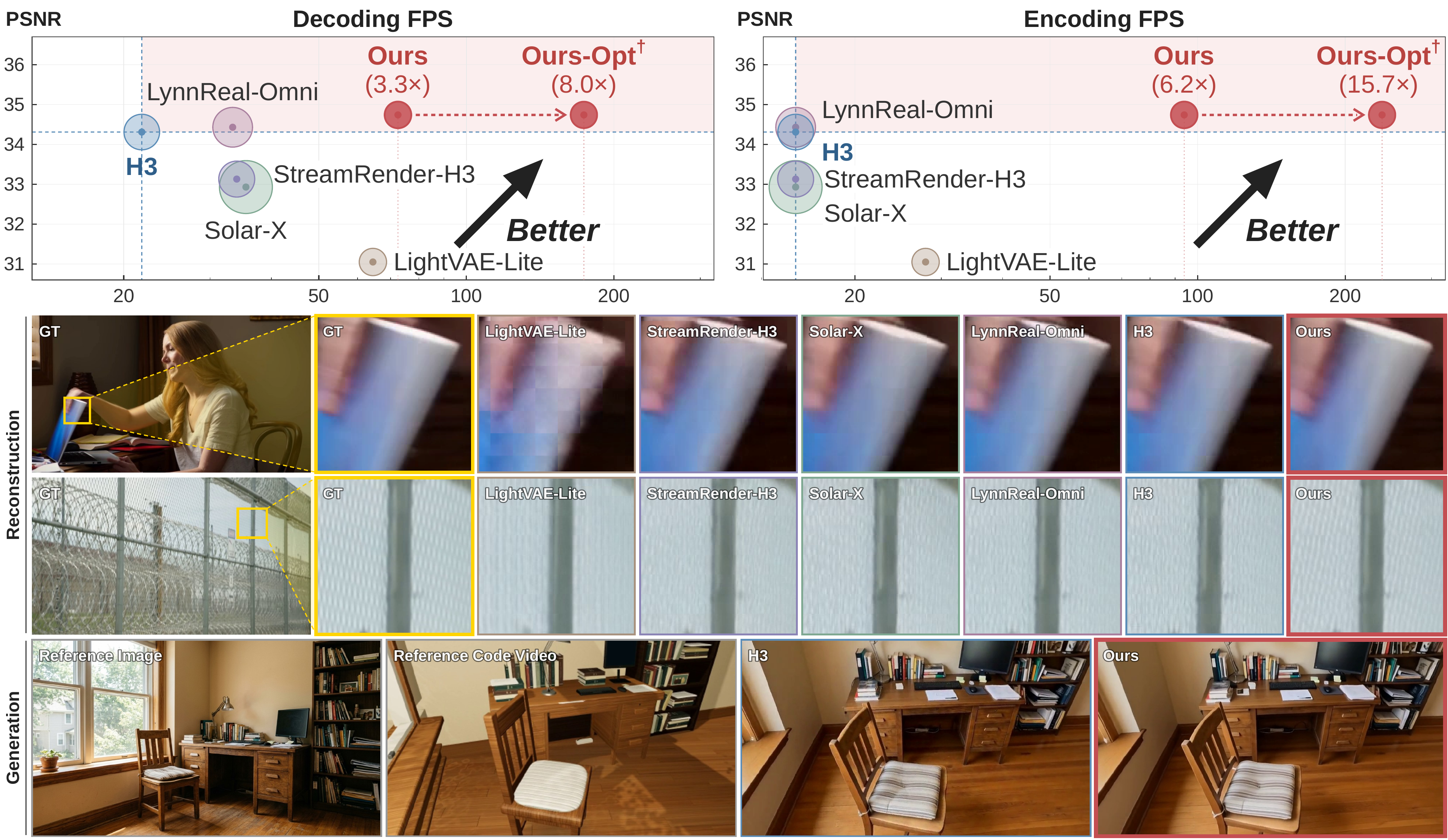}
\caption{\textbf{Speed and Quality on MiniMax-H3.} \emph{Top:} LiveVAE compared with other methods in PSNR and decoding and encoding FPS at 720p. \emph{Reconstruction Quality:} zoomed crops of the yellow box from each VAE. \emph{Generation Quality:} Reference to video results on CodeVideoBench~\citep{codevideomodel2026} with the H3 VAE and ours.}
\label{fig:teaser}
\end{figure}

Video diffusion models~\citep{rombach2022ldm,yang2025cogvideox,wan2025wan,kong2024hunyuanvideo} increasingly target real time generation, streaming video editing, and reference to video generation. Such models comprise two core components: a video VAE~\citep{kingma2014vae} that maps between pixels and a compact latent space, and a diffusion Transformer~\citep{peebles2023dit} that generates within this latent space. Recent advances in causal diffusion with KV caching~\citep{chen2024diffusionforcing,yin2025causvid} and few step autoregressive distillation~\citep{huang2025selfforcing,liu2026rollingforcing} have reduced DiT sampling to just a few causal steps, shifting the primary latency bottleneck to the video VAE. In real time generation, the Wan2.2 VAE decoder alone accounts for 64.6\% of total inference time at 720p~\citep{shin2026motionstream,kang2026flashdecoder}. Moreover, in streaming video editing and reference to video generation, the encoder must compress incoming video frames or visual references into latents on the fly, a step that can take even longer than decoding~\citep{zhao2026sanastreaming}. Supporting these interactive pipelines therefore requires fast streaming for both encoding and decoding, rather than decoding alone.

Existing video VAEs and acceleration methods struggle to meet these requirements. Mainstream video VAEs rely on heavy causal 3D convolutions~\citep{yu2024magvitv2,yang2025cogvideox,wan2025wan,kong2024hunyuanvideo} whose computational cost scales rapidly with resolution, and prior acceleration efforts focus almost exclusively on the decoder while leaving streaming encoding largely unexplored. Among decoder only methods, distilling or pruning convolutional architectures~\citep{BoerBohan2025TAEHV,gu2026lightvae,zou2025turbovaed,zhu2026flashvaed} often trades reconstruction fidelity for speed or retains substantial convolutional overhead. Streaming Transformer decoders, \eg FlashDecoder~\citep{kang2026flashdecoder}, bound temporal history with a KV cache but still attend over the entire latent, incurring quadratic cost in the number of spatial tokens. This raises a fundamental question: \textit{Do video VAE encoding and decoding actually require such global spatiotemporal modeling?}

\begin{figure}[t]
\centering
\includegraphics[width=\linewidth]{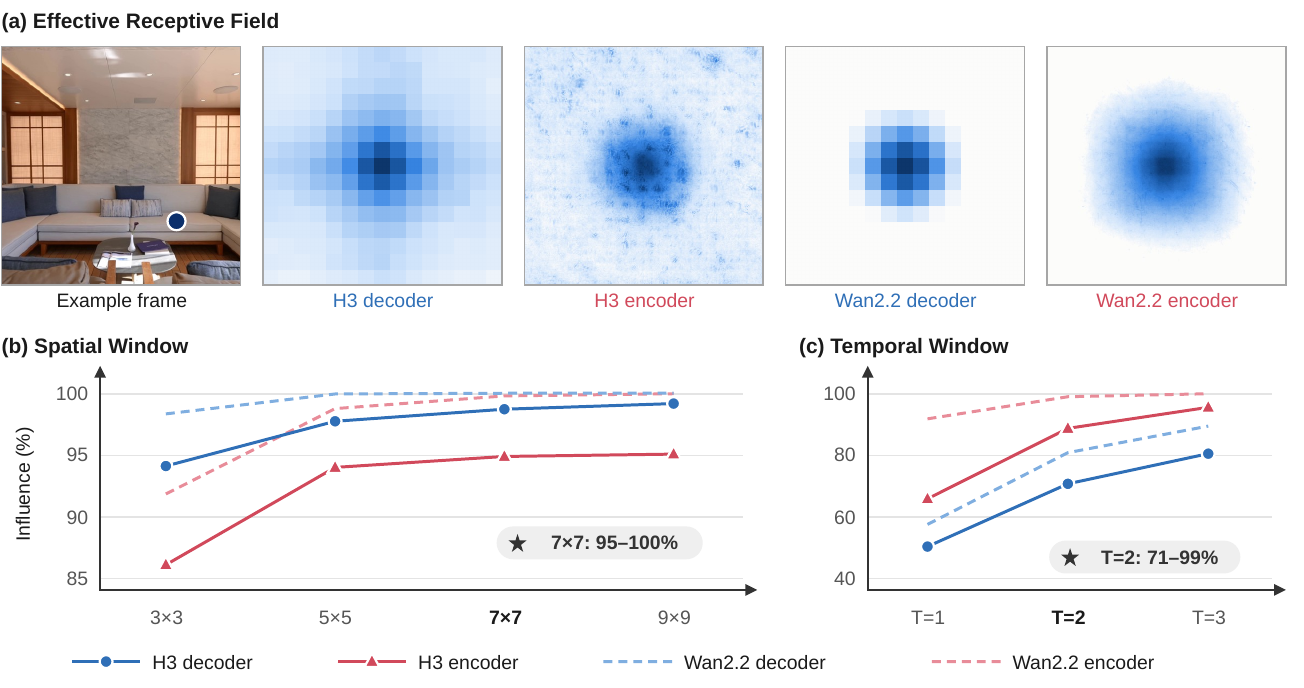}
\caption{\textbf{Effective Receptive Field analysis} of Pretrained VAEs.}
\label{fig:motivation}
\end{figure}

To answer this question, we analyze the effective receptive fields~\citep{luo2016erf} of pretrained video VAEs, including MiniMax-H3~\citep{minimax2026h3} and Wan2.2~\citep{wan2025wan}, using normalized squared input gradients, and cross check the temporal behavior with a gradient free frame replacement test~\citep{zeiler2014visualizing,huang2018makes}. As shown in \cref{fig:motivation}, the dependencies between input and output concentrate around the target in both encoders and decoders. Spatially, a compact $7\times7$ neighborhood around the target location holds 95--100\% of the gradient energy in \cref{fig:motivation}b. Temporally, the current and immediately preceding steps hold 71--99\% in \cref{fig:motivation}c. Replacing a single input step changes the output two steps later by only 0.5--1.5\% for the decoders and 4--12\% for the encoders, relative to the change caused by replacing the entire sequence.

To exploit this locality, we introduce \textbf{LiveVAE}, a streaming video encoder and decoder built on causal window attention. Spatially, restricting attention to a local window eliminates redundant computation to lower latency and enables direct extrapolation to higher resolutions by modeling only local patterns. Temporally, the causal window bounds the KV cache for continuous streaming. To preserve the pretrained latent space, we train LiveVAE via a three stage distillation framework. We first independently distill the encoder and decoder in latent and pixel spaces against the frozen pretrained VAE, and then jointly finetune both modules end to end. This alignment allows LiveVAE to directly replace the original VAE. On MiniMax-H3~\citep{minimax2026h3}, LiveVAE achieves $5.2$--$6.2\times$ faster encoding and $3.0$--$3.3\times$ faster decoding than the official VAE while matching or exceeding its PSNR from 480p to 2160p.

To further translate this architectural efficiency into hardware speedups, we build a dedicated inference infrastructure (\cref{sec:infra}) incorporating attention operator fusion, region wise compilation, and FP8 quantization aware training. With it, LiveVAE pushes the end to end speedup up to $13.4\times$ with negligible quality loss, matching the official VAE in downstream reference to video generation.

In summary, our contributions are as follows:
\begin{itemize}[nolistsep,leftmargin=*,itemsep=3pt,parsep=0pt]
\item Through effective receptive field and perturbation analyses of pretrained video VAEs, we show that video encoding and decoding rely almost entirely on local spatial and temporal neighborhoods.
\item We propose LiveVAE, the first Transformer based streaming video encoder and decoder built on causal window attention, which eliminates redundant computation to lower latency, extrapolates from 720p to 2160p with even higher PSNR, and bounds the temporal KV cache.
\item We design a three stage distillation framework that preserves the pretrained latent space, enabling LiveVAE to directly replace the original VAE.
\item We develop a dedicated inference infrastructure for LiveVAE, pushing end to end speedup up to $13.4\times$ on MiniMax-H3 with almost no quality loss.
\end{itemize}

\section{Related Work}
\label{sec:related}

\paragraph{Latent Diffusion Models.}
Latent diffusion~\citep{rombach2022ldm} trains the generator in the compact latent space of a pretrained VAE~\citep{kingma2014vae}, and diffusion Transformers (DiTs)~\citep{peebles2023dit} have scaled this recipe to video~\citep{yang2025cogvideox,wan2025wan,kong2024hunyuanvideo,hacohen2024ltx,nvidia2025cosmos}. The latent space is the representation the DiT learns from, and the quality of this representation largely determines how fast the DiT converges and how well it generates~\citep{yu2025repa,singh2026irepa,yao2025vavae}. Existing work therefore designs the latent space for the generator. One line raises the compression ratio so that the DiT processes fewer tokens~\citep{chen2025dcae,hacohen2024ltx}, with causal 3D convolutions that compress time as well~\citep{yu2024magvitv2,yang2025cogvideox}. Another aligns the latents with pretrained visual representations. VA-VAE~\citep{yao2025vavae} aligns them with vision foundation models to ease the tradeoff between reconstruction and generation, and REPA-E~\citep{leng2025repae} tunes the VAE and the DiT end to end through such an alignment. A third uses these representations directly as latents. RAE~\citep{zheng2026rae} pairs a frozen representation encoder with a trained decoder. Each design defines a new latent space on which the DiT is then trained, and once the DiT is trained, this latent space is fixed. CV-VAE~\citep{zhao2024cvvae} thus keeps a new video VAE compatible with the latent space of a pretrained image VAE. Our work is orthogonal to these designs. Whichever latent space a DiT is trained on, we keep it and replace only the encoder and decoder that compute it with streaming ones.

\paragraph{Streaming Video Generation.}
On the generator side, video generation has become causal and fast. Diffusion Forcing~\citep{chen2024diffusionforcing} and CausVid~\citep{yin2025causvid} generate one frame at a time with a KV cache, and Self Forcing~\citep{huang2025selfforcing} and Rolling Forcing~\citep{liu2026rollingforcing} close the train-test gap and generate long videos in real time with a few denoising steps. Once the generator needs only a few causal steps, the bottleneck shifts to the VAE. Few step autoregressive distillation raises the decoder's share of latency by more than ten times~\citep{zhu2026flashvaed}, the Wan2.2 decoder takes 64.6\% of the inference time of MotionStream~\citep{shin2026motionstream} at 720p~\citep{kang2026flashdecoder}, and in streaming video editing, where the input video is also encoded, VAE encoding takes longer than decoding~\citep{zhao2026sanastreaming}. Streaming generation thus needs both a streaming decoder and a streaming encoder.

\paragraph{Video VAE Acceleration.}
Almost all video VAE acceleration methods accelerate only the decoder. We group them by whether they keep the latent space of the DiT. Lightweight VAEs trained from scratch~\citep{cheng2025leanvae,li2025wfvae,wu2025h3ae} define a new latent space, so the DiT must be retrained or finetuned. The others keep the latent space. Distilled and pruned decoders reduce per frame cost. TAEHV~\citep{BoerBohan2025TAEHV} and LightVAE~\citep{gu2026lightvae} distill a lighter decoder that trades fidelity for speed, while Turbo-VAED~\citep{zou2025turbovaed} and Flash-VAED~\citep{zhu2026flashvaed} prune and replace the original decoder's causal 3D convolutions, whose cost still grows with resolution. Streaming decoders instead reduce latency by decoding as the latents arrive. Some make a bidirectional decoder causal~\citep{chen2026h3streaming,streamrenderh3_2026,zhao2026sanastreaming}, and FlashDecoder~\citep{kang2026flashdecoder}, closest to ours, is a Transformer decoder that streams one latent frame at a time with a rolling KV cache. Its tokens attend to the full current and previous frames, so its cost grows quadratically with resolution, and it requires multiresolution training up to higher resolution; ours attend to a fixed local window, scale linearly, and are trained only at low resolutions, yet extrapolate directly to higher resolutions at inference time.

\section{Method}
\label{sec:method}

\begin{figure*}[t]
\centering
\includegraphics[width=\linewidth]{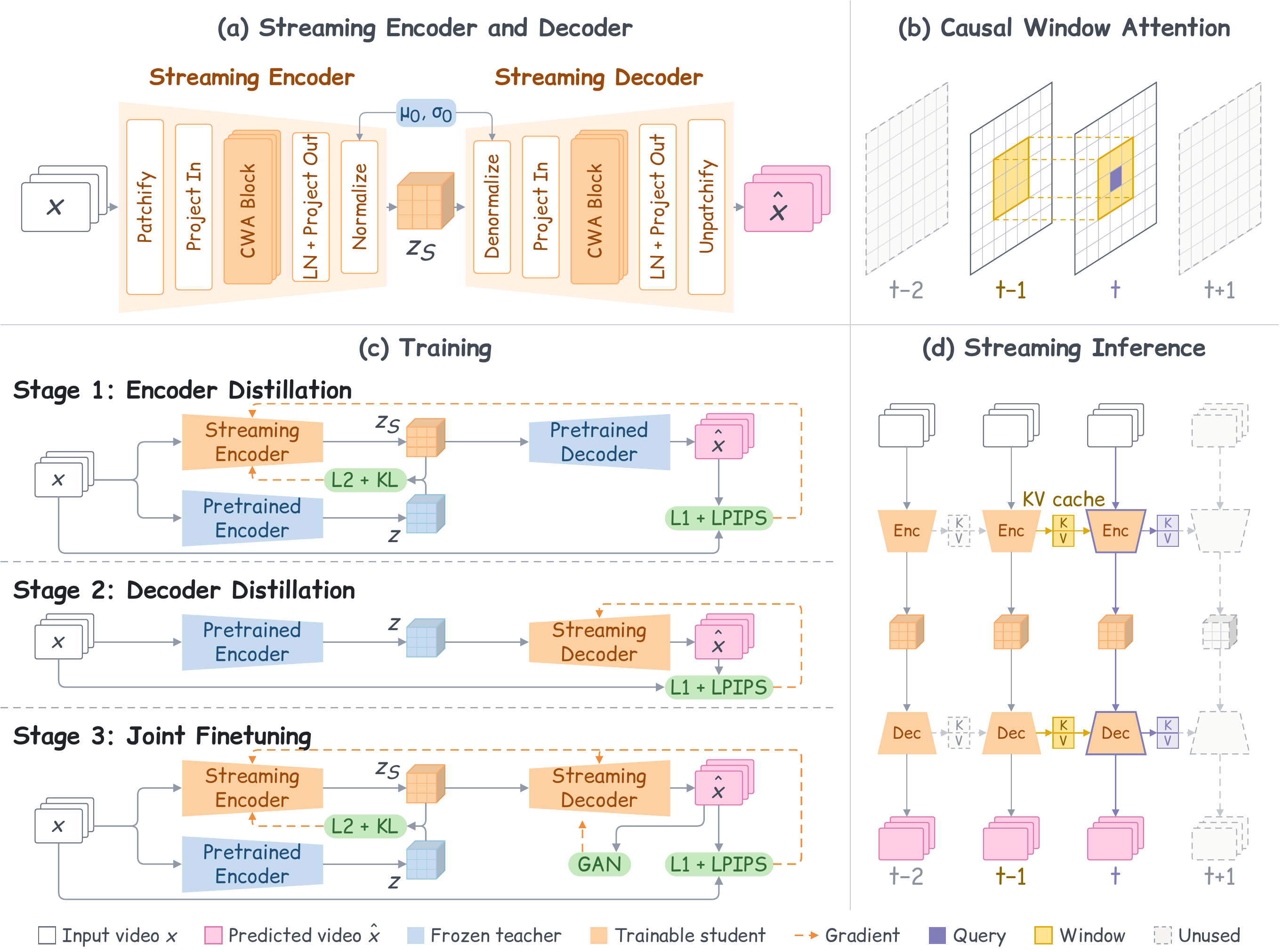}
\caption{\textbf{Overview of LiveVAE.} (a) Encoder and Decoder architecture of LiveVAE. (b) Causal Window Attention. (c) Three stage progressive distillation. (d) Streaming Inference with a bounded KV cache.}
\label{fig:method}
\end{figure*}

We distill pretrained video VAEs into \textbf{LiveVAE}, a streaming autoencoder built on causal window attention that preserves full compatibility with existing diffusion models as illustrated in \cref{fig:method}. \cref{sec:prelim} reviews the preliminaries and problem formulation, \cref{sec:na} introduces causal window attention, \cref{sec:distill} presents our network architecture and three stage progressive distillation, and \cref{sec:infra} details the inference infrastructure optimization.

\subsection{Preliminaries}
\label{sec:prelim}

A pretrained video VAE consists of a teacher encoder $\mathcal{E}_T$ and decoder $\mathcal{D}_T$ with temporal and spatial compression ratios $M$ and $S$. Given an input video $x\in\mathbb{R}^{Mf\times Sh\times Sw\times3}$, the encoder maps $x$ to a latent representation $z\in\mathbb{R}^{f\times h\times w\times C}$, and the decoder reconstructs $\hat{x}$ from $z$. Along the temporal axis, $x$ divides into $f$ consecutive frame chunks $\{x_t\}_{t=1}^{f}$ and $z$ into $f$ latent frames $\{z_t\}_{t=1}^{f}$, where the $t$-th frame chunk and the $t$-th latent frame are given by
\begin{equation}
x_t\in\mathbb{R}^{M\times Sh\times Sw\times 3}\qquad z_t\in\mathbb{R}^{h\times w\times C}
\label{eq:chunk}
\end{equation}
Our streaming encoder $\mathcal{E}_S$ and decoder $\mathcal{D}_S$ replace the teacher modules and operate directly on tokens of the $f\times h\times w$ latent grid, where token $i=(t,y,x)$ lies in latent frame $t\in\{1,\dots,f\}$ at spatial location $(y,x)$.

\subsection{Causal Window Attention}
\label{sec:na}

\paragraph{3D Causal Window.}
Motivated by the strong locality of video VAEs in \cref{fig:motivation}, causal window attention, denoted CWA, restricts each token to a local $H\times W$ spatial neighborhood across the current and preceding $T-1$ latent frames as shown in \cref{fig:method}b. For a token $i=(t,y,x)$, its causal window $\mathcal{W}(i)$ contains the tokens $(\tau,y',x')$ satisfying
\begin{equation}
0\le t-\tau<T,\quad |y'-y|\le\tfrac{H-1}{2},\quad |x'-x|\le\tfrac{W-1}{2}
\label{eq:window}
\end{equation}
for interior positions with odd $H$ and $W$. Along spatial borders, the window shifts inward instead of zero padding to retain $H\times W$ tokens per latent frame~\citep{hassani2023nat}. Let $q_i,k_i,v_i\in\mathbb{R}^d$ be the query, key, and value of token $i$, after rotating queries and keys with 3D RoPE aligned with the base DiT in \cref{sec:encdec}. Over this window of at most $THW$ tokens, CWA computes
\begin{equation}
\mathrm{CWA}(i)=\sum_{j\in\mathcal{W}(i)}\frac{\exp(q_i k_j^{\top}/\sqrt{d})}{\sum_{m\in\mathcal{W}(i)}\exp(q_i k_m^{\top}/\sqrt{d})}\,v_j
\label{eq:cwa}
\end{equation}

\paragraph{Frame Wise 2D Decomposition.}
To compute \cref{eq:cwa} with optimized 2D neighborhood attention kernels~\citep{hassani2023nat} rather than irregular 3D memory accesses, we partition $\mathcal{W}(i)$ by latent frame and denote by $\mathcal{W}_\tau(i)$ the $H\times W$ spatial neighbors of token $i$ in latent frame $\tau$. Applying 2D window attention on latent frame $\tau$ yields a per frame output $O_\tau(i)\in\mathbb{R}^d$ and its log-sum-exp normalizer $L_\tau(i)\in\mathbb{R}$:
\begin{equation}
\begin{aligned}
O_\tau(i)&=\sum_{j\in\mathcal{W}_\tau(i)}\frac{\exp(q_i k_j^{\top}/\sqrt{d})}{\exp\bigl(L_\tau(i)\bigr)}\,v_j\\
L_\tau(i)&=\log\sum_{j\in\mathcal{W}_\tau(i)}\exp(q_i k_j^{\top}/\sqrt{d})
\end{aligned}
\label{eq:per_frame}
\end{equation}
We then recover the exact 3D attention output by weighting the per frame outputs with normalizers rescaled across the $T$ latent frames:
\begin{equation}
\begin{aligned}
\mathrm{CWA}(i)&=\sum_{\tau=t-T+1}^{t}\alpha_\tau(i)\,O_\tau(i)\\
\alpha_\tau(i)&=\frac{\exp\bigl(L_\tau(i)\bigr)}{\sum_{\tau'=t-T+1}^{t}\exp\bigl(L_{\tau'}(i)\bigr)}
\end{aligned}
\label{eq:merge}
\end{equation}

\paragraph{Streaming Inference with KV Cache.}
\label{sec:stream}
During streaming inference, the models process one latent frame $t$ at a time as shown in \cref{fig:method}d: the encoder maps incoming frame chunk $x_t$ to latent frame $z_t$, and the decoder reconstructs $\hat{x}_t$ from $z_t$. Each CWA block maintains a rolling cache storing the key value pairs $\{(K_\tau,V_\tau)\}_{\tau=t-T+1}^{t-1}$ from the preceding $T-1$ latent frames. For latent frame $t$, the block computes only $Q_t,K_t,V_t\in\mathbb{R}^{h\times w\times d}$, pairs $Q_t$ with the cached keys and values for $\tau < t$ and the newly computed $(K_t,V_t)$ to evaluate \cref{eq:per_frame,eq:merge}, and then updates the cache by appending $(K_t,V_t)$ and evicting $(K_{t-T+1},V_{t-T+1})$. Across $n=fhw$ tokens, CWA requires $\mathcal{O}(nTHW)$ operations, and the cache size remains fixed at $T-1$ latent frames regardless of the input video duration.

\subsection{Architecture and Progressive Distillation}
\label{sec:encdec}
\label{sec:distill}
\label{sec:method_encoder}

Unlike multiscale convolutional VAEs that expend heavy computation on high resolution feature maps, $\mathcal{E}_S$ and $\mathcal{D}_S$ use a symmetric backbone of $L$ CWA blocks operating directly on the $f\times h\times w$ latent grid in \cref{fig:method}a. The encoder patchifies incoming frame chunks via a linear projection, processes tokens through the CWA blocks, and projects them to the latent channels before normalizing with the teacher's channelwise statistics; symmetrically, the decoder denormalizes incoming latents, applies the CWA blocks, and unpatchifies hidden features back to pixels. 

We list the layers of both networks below in the order in which they are applied, with the hidden dimension $D$ given in \cref{sec:setup}.

\paragraph{Encoder.}
The encoder $\mathcal{E}_S$ maps the video $x$ to the parameters of the latent posterior in five steps.
\begin{enumerate}[nolistsep,leftmargin=*,itemsep=3pt,parsep=0pt]
\item \textbf{Patch embedding.} A space to depth operator rearranges each $M\times S\times S$ block of pixels into one vector of $3MS^2$ values, which turns every frame chunk $x_t$ into $h\times w$ patches. The patches do not overlap and never mix frames of two frame chunks, so the encoder consumes the video chunk by chunk.
\item \textbf{Project in.} A linear layer maps every patch from $3MS^2$ to $D$ channels and yields the tokens $\mathbf{u}\in\mathbb{R}^{f\times h\times w\times D}$.
\item \textbf{CWA blocks.} The tokens pass through $L$ blocks. Each block applies layer normalization before CWA and before the FFN, with a residual connection around both:
\begin{equation}
\begin{aligned}
\mathbf{u}'&=\mathbf{u}+\mathrm{CWA}\bigl(\mathrm{LN}(\mathbf{u})\bigr)\\
\mathbf{u}&\leftarrow\mathbf{u}'+\mathrm{FFN}\bigl(\mathrm{LN}(\mathbf{u}')\bigr)
\end{aligned}
\label{eq:block}
\end{equation}
All $L$ blocks share the same $T\times H\times W$ window and the same hidden dimension $D$, with attention heads of dimension $d=64$. Queries and keys are normalized and then rotated by the 3D RoPE, which follows the specification of the DiT of the base model, with the same split of the head dimension over the three axes and the same base frequencies (\cref{sec:abl_rope}).
\item \textbf{Project out.} A layer normalization and a linear layer map every token from $D$ to $2C$ channels, the raw posterior mean $\mu_{\mathrm{raw}}$ and log variance $\log\sigma_{\mathrm{raw}}^2$.
\item \textbf{Normalization.} The DiT is trained on latents standardized by the channelwise statistics $\mu_0,\sigma_0\in\mathbb{R}^C$ of the base VAE, so the encoder applies the same standardization as its last step:
\begin{equation}
z_S=\frac{\mu_{\mathrm{raw}}-\mu_0}{\sigma_0}\qquad
\log\sigma_S^2=\log\sigma^2_{\mathrm{raw}}-2\log\sigma_0
\label{eq:moments}
\end{equation}
The output $z_S$ thus lies in the latent space on which the DiT is trained and needs no further scaling.
\end{enumerate}

\paragraph{Decoder.}
The decoder $\mathcal{D}_S$ mirrors the encoder and maps a standardized latent $z$, produced by $\mathcal{E}_S$ or by the DiT, back to video frames in five steps.
\begin{enumerate}[nolistsep,leftmargin=*,itemsep=3pt,parsep=0pt]
\item \textbf{Denormalization.} The decoder first restores the native latent scale of the base VAE, $\sigma_0\odot z+\mu_0$, the inverse of the last step of the encoder.
\item \textbf{Project in.} A linear layer maps every latent token from $C$ to $D$ channels and yields the tokens $\mathbf{u}\in\mathbb{R}^{f\times h\times w\times D}$.
\item \textbf{CWA blocks.} The tokens pass through $L$ blocks of the form of \cref{eq:block}, with the same window, head dimension and RoPE as in the encoder.
\item \textbf{Project out.} A layer normalization and a linear layer map every token from $D$ to $3MS^2$ channels, the pixel values of one patch.
\item \textbf{Unpatchify.} A depth to space operator rearranges each vector of $3MS^2$ values into an $M\times S\times S$ block of pixels, which gives the frame chunk $\hat{x}_t$ of every latent frame $z_t$.
\end{enumerate}

To preserve the pretrained teacher latent space, we introduce a three stage progressive distillation framework in \cref{fig:method}c that first distills each student module independently against the frozen teacher and then finetunes both modules jointly. Across all stages, pixel reconstruction is supervised by an $\ell_1$ loss and perceptual LPIPS~\citep{zhang2018lpips}:
\begin{equation}
\mathcal{L}_{\mathrm{rec}}(x,\hat{x})=\|x-\hat{x}\|_1+\lambda_{\mathrm{p}}\,\mathrm{LPIPS}(x,\hat{x})
\label{eq:rec}
\end{equation}

\paragraph{Stage 1: Encoder Distillation.}
Let $z_S=\mathcal{E}_S(x)$ and $z=\mathcal{E}_T(x)$ denote the student and teacher latents, with corresponding posteriors $q_S(z|x)$ and $q_T(z|x)$. We train $\mathcal{E}_S$ to match the teacher in latent space while ensuring $z_S$ remains decodable by the frozen teacher decoder $\mathcal{D}_T$:
\begin{equation}
\mathcal{L}_{\mathcal{E}}=\mathcal{L}_{\mathrm{lat}}+\mathcal{L}_{\mathrm{rec}}\bigl(x,\mathcal{D}_T(z_S)\bigr)
\label{eq:enc}
\end{equation}
where $\mathcal{L}_{\mathrm{lat}}$ combines an $\ell_2$ loss with posterior KL divergence:
\begin{equation}
\mathcal{L}_{\mathrm{lat}}=\lambda_z\|z_S-z\|_2^2+\lambda_{\mathrm{KL}}\,D_{\mathrm{KL}}\bigl(q_S(z|x)\,\|\,q_T(z|x)\bigr)
\label{eq:kl}
\end{equation}
Here, $\mathcal{L}_{\mathrm{lat}}$ enforces explicit alignment in latent space, whereas backpropagating $\mathcal{L}_{\mathrm{rec}}$ through the frozen $\mathcal{D}_T$ penalizes latent deviations that degrade visual fidelity.

\paragraph{Stage 2: Decoder Distillation.}
In parallel, we train $\mathcal{D}_S$ by encoding input $x$ with the frozen teacher encoder $\mathcal{E}_T$, decoding it with $\mathcal{D}_S$, and computing the reconstruction loss against $x$:
\begin{equation}
\mathcal{L}_{\mathcal{D}}=\mathcal{L}_{\mathrm{rec}}\bigl(x,\mathcal{D}_S(\mathcal{E}_T(x))\bigr)
\label{eq:dec}
\end{equation}

\paragraph{Stage 3: Joint Finetuning.}
Finally, we finetune $\mathcal{E}_S$ and $\mathcal{D}_S$ end to end so that $\mathcal{D}_S$ adapts to student latents $z_S=\mathcal{E}_S(x)$ while $\mathcal{E}_S$ remains anchored to the teacher's latent space:
\begin{equation}
\mathcal{L}_{\mathrm{joint}}=\mathcal{L}_{\mathrm{lat}}+\mathcal{L}_{\mathrm{rec}}\bigl(x,\mathcal{D}_S(z_S)\bigr)+w\,\mathcal{L}_G\bigl(\mathcal{D}_S(z_S)\bigr)
\label{eq:joint}
\end{equation}
where $\mathcal{L}_G$ is an adversarial loss computed with a pretrained Real-ESRGAN U-Net discriminator~\citep{wang2021realesrgan} to sharpen high frequency details. Following \citet{esser2021taming}, the adaptive weight $w$ caps the adversarial gradient norm at $\lambda_{\mathrm{adv}}$ times the reconstruction gradient norm. During training, we keep an exponential moving average of the model weights~\citep{polyak1992averaging,karras2018progressive}.

\paragraph{Latent Alignment.}
In $\mathcal{L}_{\mathrm{lat}}$ of \cref{eq:kl}, the $\ell_2$ term aligns the student and teacher latents. With diagonal Gaussian posteriors $q_T(z|x)=\mathcal{N}(z;\mathcal{E}_T(x),\sigma_T^2)$ and $q_S(z|x)=\mathcal{N}(z;z_S,\sigma_S^2)$, the KL term additionally weights mean errors by the teacher precision $1/\sigma_T^2$ and aligns $\sigma_S$ with $\sigma_T$, which supervises the log variance output of the student encoder. We floor $\sigma_T$ so that very small teacher variances do not dominate the loss.

\paragraph{Data and Supervision.}
All three stages are trained on the same mix of images and videos (\cref{sec:setup}), and the reconstruction loss always compares the output with the input $x$ rather than with a teacher reconstruction.

\paragraph{Adversarial Training.}
The discriminator $\Phi$ is initialized from the pretrained Real-ESRGAN U-Net weights, produces a per pixel logit map and is applied frame by frame. In our runs, training a discriminator from scratch did not reliably distinguish removing high frequencies from restoring them, whereas the pretrained discriminator scores blurred frames as fake. With $\hat{x}=\mathcal{D}_S(z_S)$, we use the non-saturating losses
\begin{equation}
\begin{aligned}
\mathcal{L}_{\Phi}&=\mathrm{BCE}\bigl(\Phi(x),1\bigr)+\mathrm{BCE}\bigl(\Phi(\mathrm{sg}[\hat{x}]),0\bigr)\\
\mathcal{L}_G&=-\log\mathrm{sigmoid}\bigl(\Phi(\hat{x})\bigr)
\end{aligned}
\label{eq:gan}
\end{equation}
where $\mathrm{sg}$ stops the gradient, and set the adaptive weight of \cref{eq:joint} as
\begin{equation}
w=\lambda_{\mathrm{adv}}\min\!\left(\frac{\|\nabla_{\hat{x}}\mathcal{L}_{\mathrm{rec}}\|}{\|\nabla_{\hat{x}}\mathcal{L}_G\|},\,1\right)
\label{eq:adv}
\end{equation}
In our runs with a fixed weight, the adversarial term tended to dominate and training became unstable within a few hundred steps. The EMA decay and the optimization settings are given in \cref{sec:setup}.

\subsection{Inference Infrastructure Optimization}
\label{sec:infra}

To further accelerate streaming inference, we introduce three system optimizations on NVIDIA H100 GPUs rented from RunPod that yield up to a $3.0\times$ end to end speedup with negligible quality loss (\cref{fig:infra}). These optimizations are enabled only for Ours-Opt in \cref{tab:main}.

\paragraph{Attention Operator Fusion.}
Running query key normalization, 3D RoPE, 2D window attention, and the cross frame merge of \cref{eq:merge} separately incurs heavy memory traffic from intermediate tensors. We fuse query key normalization with 3D RoPE into one kernel and fold the cross frame merge directly into the window attention kernel, so that the fused attention path reads and writes global memory only at its inputs and outputs. The window attention kernel attends over the full causal window with an online softmax, so the merge needs no separate pass.

\paragraph{Region Wise Compilation.}
Compiling the whole model with \texttt{torch.compile} suffers from graph breaks around custom attention kernels and recompilations during KV cache warmup. Instead, we compile only the four static subgraphs in each block, namely the normalization before attention, QKV projection, output projection, and the residual FFN, while leaving the CWA kernels and cache control flow uncompiled in the outer loop.

\paragraph{FFN FP8 with QAT.}
With attention made lightweight, the FFN dominates computation, so we quantize the FFN linear layers to FP8 in all encoder blocks and in every decoder block except the last. To avoid the quality drop of direct post-training quantization, we apply quantization aware training, denoted QAT, under \cref{eq:joint} by simulating FP8 precision during finetuning. Attention and normalization retain their original precision.

\emph{Choice of layers.}
We enable FP8 in one block at a time on the model before QAT and measure the change in reconstruction PSNR. The cost is concentrated in the last block of each network: at 480p, PSNR drops by about 0.9\,dB for the last encoder block and by more than 7\,dB for the last decoder block, against about 0.1\,dB for the other 22 blocks together. The inputs of the last decoder FFN far exceed the range of FP8, so part of them saturates. This loss appears to come from dynamic range rather than from rounding precision, which we expect finetuning to recover less easily, while leaving one block out costs little speed. We therefore quantize the FFN of all 12 encoder blocks and of the first 11 decoder blocks, and keep the last decoder block in its original precision. The last encoder block stays quantized, and its loss is the main part that QAT has to recover.

\emph{QAT training.}
QAT starts from the model after Stage~3 and continues training with the quantized forward pass, so that the weights adapt to the deterministic error of FP8. It rests on the following four choices.
\begin{itemize}[nolistsep,leftmargin=*,itemsep=3pt,parsep=0pt]
\item \textbf{Deployed forward pass, straight through gradients.} The forward pass simulates FP8 by calling the same FP8 multiplication as the inference engine, so training and deployment see the same values. For a quantized linear layer with input $a$ and weight $\Theta$,
\begin{equation}
\tilde{y}=\mathrm{FP8}(a,\Theta)+\bigl(a\Theta^\top-\mathrm{sg}[a\Theta^\top]\bigr)
\label{eq:ste}
\end{equation}
where $\mathrm{FP8}(a,\Theta)$ is the deployed FP8 product. The bracket is zero in the forward pass, so $\tilde{y}$ takes the FP8 value, while the gradients are those of the ordinary linear layer, as in a straight through estimator.
\item \textbf{Floating point weights.} Only the quantized layers are updated, and the optimizer always updates their floating point weights. The quantized forward pass adds no parameter and no running statistic, so the checkpoint keeps the format of the unquantized model and is converted to FP8 only once, at deployment time.
\item \textbf{Unchanged objective.} The objective, the loss weights and the data mix follow Stage~3, so that the change is limited to FP8. The latent alignment terms keep the encoder in the latent space of the teacher while its weights adapt to the quantization error of FP8.
\item \textbf{Delayed adversarial term.} Because the optimizer state is reinitialized, the adversarial term of the generator is enabled only after a short warmup, during which the discriminator, continued from Stage~3, is already updated.
\end{itemize}
We export the EMA weights. They are specific to FP8: under FP8 they bring PSNR back to within 0.13\,dB of the unquantized model (\cref{tab:infra_full}), while in the original precision they fall behind the original weights, so we use them only when FP8 is enabled. These three optimizations together define Ours-Opt, with FP8 applied only at 1080p and 2160p in \cref{tab:main}, whereas Ours, the row compared with the baselines, runs without any of them.

\newcommand{\tms}[2]{#1\,\makebox[3.1em][l]{\textcolor{red}{{\fontsize{6}{7}\selectfont\bfseries\boldmath #2$\times$}}}}
\newcommand{\tmo}[1]{#1\,\makebox[3.1em][l]{}}
\begin{table*}[t]
\centering
\footnotesize
\setlength{\tabcolsep}{3pt}
\setlength{\aboverulesep}{0pt}\setlength{\belowrulesep}{0pt}\renewcommand{\arraystretch}{1.02}
\caption{Reconstruction quality and efficiency. An asterisk (*) marks a streaming encoder or decoder; a single model name denotes both components. Red: speedup over the official teacher VAE. Bold: best quality metric excluding the official teacher VAE. Ours-Opt adds the inference optimizations of \cref{sec:infra}, with FP8 applied at 1080p and 2160p.}
\vspace{-0.8em}
\label{tab:main}\label{tab:efficiency}\label{tab:full}
\resizebox{\textwidth}{!}{%
\begin{tabular}{l c ccc rrr ccc rrr}
\toprule
 & & \multicolumn{6}{c}{\textbf{480p}} & \multicolumn{6}{c}{\textbf{720p}} \\
\cmidrule(lr){3-8} \cmidrule(lr){9-14}
\multicolumn{1}{c}{\begin{tabular}[c]{@{}c@{}}\textbf{Encoder / Decoder}\end{tabular}} & \multicolumn{1}{c}{\begin{tabular}[c]{@{}c@{}}\textbf{Params}\end{tabular}} & \multicolumn{1}{c}{\begin{tabular}[c]{@{}c@{}}\textbf{PSNR}\end{tabular}} & \multicolumn{1}{c}{\begin{tabular}[c]{@{}c@{}}\textbf{SSIM}\end{tabular}} & \multicolumn{1}{c}{\begin{tabular}[c]{@{}c@{}}\textbf{LPIPS}\end{tabular}} & \multicolumn{1}{c}{\begin{tabular}[c]{@{}c@{}}\textbf{Enc. Time}\end{tabular}} & \multicolumn{1}{c}{\begin{tabular}[c]{@{}c@{}}\textbf{Dec. Time}\end{tabular}} & \multicolumn{1}{c}{\begin{tabular}[c]{@{}c@{}}\textbf{Mem}\end{tabular}} & \multicolumn{1}{c}{\begin{tabular}[c]{@{}c@{}}\textbf{PSNR}\end{tabular}} & \multicolumn{1}{c}{\begin{tabular}[c]{@{}c@{}}\textbf{SSIM}\end{tabular}} & \multicolumn{1}{c}{\begin{tabular}[c]{@{}c@{}}\textbf{LPIPS}\end{tabular}} & \multicolumn{1}{c}{\begin{tabular}[c]{@{}c@{}}\textbf{Enc. Time}\end{tabular}} & \multicolumn{1}{c}{\begin{tabular}[c]{@{}c@{}}\textbf{Dec. Time}\end{tabular}} & \multicolumn{1}{c}{\begin{tabular}[c]{@{}c@{}}\textbf{Mem}\end{tabular}} \\
\midrule
\rowcolor{gray!15}
Wan2.2\textsuperscript{*} & 0.70\,B & 36.08 & .945 & .019 & \tmo{175.0} & \tmo{560.3} & 11.0\,GB & 37.66 & .954 & .018 & \tmo{382.8} & \tmo{1264.2} & 21.8\,GB \\
TAEW2.2\textsuperscript{*} & 0.01\,B & 30.10 & .857 & .071 & \tms{14.4}{12.2} & \tms{16.9}{33.2} & 0.1\,GB & 31.49 & .883 & .068 & \tms{17.9}{21.4} & \tms{22.8}{55.4} & 0.3\,GB \\
TAEW2.2-Super\textsuperscript{*} & 0.02\,B & 31.49 & .884 & .064 & \tms{12.1}{14.5} & \tms{22.0}{25.5} & 0.2\,GB & 32.89 & .905 & .060 & \tms{18.0}{21.3} & \tms{46.8}{27.0} & 0.3\,GB \\
LightTAE\textsuperscript{*} & 0.01\,B & 27.93 & .808 & .109 & \tms{14.3}{12.2} & \tms{16.7}{33.6} & 0.1\,GB & 28.88 & .834 & .107 & \tms{18.4}{20.8} & \tms{23.2}{54.5} & 0.3\,GB \\
Wan2.2\textsuperscript{*} / LightVAE-Lite\textsuperscript{*} & 0.30\,B & 32.64 & .898 & .069 & \tmo{175.0} & \tms{196.1}{2.9} & 4.4\,GB & 34.28 & .917 & .062 & \tmo{382.8} & \tms{438.5}{2.9} & 9.1\,GB \\
Wan2.2\textsuperscript{*} / LightVAE-Pro\textsuperscript{*} & 0.50\,B & \textbf{35.30} & \textbf{.935} & \textbf{.032} & \tmo{175.0} & \tms{358.0}{1.6} & 4.9\,GB & \textbf{36.91} & \textbf{.947} & \textbf{.031} & \tmo{382.8} & \tms{786.2}{1.6} & 9.8\,GB \\
Wan2.2\textsuperscript{*} / Turbo-VAED & 0.25\,B & 32.79 & .900 & .055 & \tmo{175.0} & \tms{306.1}{1.8} & 7.5\,GB & 34.37 & .920 & .052 & \tmo{382.8} & \tms{708.7}{1.8} & 16.4\,GB \\
\addlinespace[2pt]
\textbf{Ours}\textsuperscript{*} & 2.47\,B & 34.46 & .927 & .042 & \tms{124.9}{1.4} & \tms{119.5}{4.7} & 5.2\,GB & 36.24 & .941 & .038 & \tms{261.5}{1.5} & \tms{254.9}{5.0} & 5.9\,GB \\
\midrule
\rowcolor{gray!15}
H3 & 2.60\,B & 32.99 & .909 & .041 & \tmo{623.0} & \tmo{435.3} & 8.8\,GB & 34.31 & .923 & .038 & \tmo{1453.5} & \tmo{1010.6} & 9.0\,GB \\
TAEH3 & 0.01\,B & 27.59 & .794 & .119 & \tms{7.6}{82.0} & \tms{9.4}{46.3} & 1.4\,GB & 28.72 & .825 & .113 & \tms{15.4}{94.4} & \tms{20.3}{49.8} & 3.3\,GB \\
LightVAE-Lite & 0.91\,B & 29.71 & .845 & .127 & \tms{339.4}{1.8} & \tms{146.9}{3.0} & 5.2\,GB & 31.05 & .870 & .112 & \tms{789.5}{1.8} & \tms{341.3}{3.0} & 5.3\,GB \\
\begin{tabular}[c]{@{}l@{}}H3 / LynnReal-Omni\end{tabular} & 1.93\,B & \textbf{33.16} & \textbf{.912} & \textbf{.044} & \tmo{623.0} & \tms{285.7}{1.5} & 10.9\,GB & 34.43 & \textbf{.925} & \textbf{.040} & \tmo{1453.5} & \tms{659.5}{1.5} & 11.2\,GB \\
H3 / Solar-X\textsuperscript{*} & 2.60\,B & 31.62 & .884 & .051 & \tmo{623.0} & \tms{273.5}{1.6} & 11.3\,GB & 32.93 & .902 & .048 & \tmo{1453.5} & \tms{619.7}{1.6} & 19.9\,GB \\
\begin{tabular}[c]{@{}l@{}}H3 / StreamRender-H3\textsuperscript{*}\end{tabular} & 1.80\,B & 31.86 & .893 & .052 & \tmo{623.0} & \tms{331.4}{1.3} & 7.4\,GB & 33.13 & .909 & .049 & \tmo{1453.5} & \tms{647.0}{1.6} & 9.2\,GB \\
\addlinespace[2pt]
\textbf{Ours}\textsuperscript{*} & 2.08\,B & 33.06 & .904 & .054 & \tms{119.5}{5.2} & \tms{146.0}{3.0} & 4.4\,GB & \textbf{34.74} & .922 & .049 & \tms{234.3}{6.2} & \tms{303.4}{3.3} & 5.0\,GB \\
\textbf{Ours-Opt}\textsuperscript{*} & 2.08\,B & 33.06 & .904 & .054 & \tms{78.6}{7.9} & \tms{78.6}{5.5} & 4.4\,GB & \textbf{34.74} & .922 & .049 & \tms{92.5}{15.7} & \tms{126.7}{8.0} & 5.0\,GB \\
\specialrule{\heavyrulewidth}{0pt}{0pt}
 & & \multicolumn{6}{c}{\textbf{1080p}} & \multicolumn{6}{c}{\textbf{2160p}} \\
\cmidrule(lr){3-8} \cmidrule(lr){9-14}
\multicolumn{1}{c}{\begin{tabular}[c]{@{}c@{}}\textbf{Encoder / Decoder}\end{tabular}} & \multicolumn{1}{c}{\begin{tabular}[c]{@{}c@{}}\textbf{Params}\end{tabular}} & \multicolumn{1}{c}{\begin{tabular}[c]{@{}c@{}}\textbf{PSNR}\end{tabular}} & \multicolumn{1}{c}{\begin{tabular}[c]{@{}c@{}}\textbf{SSIM}\end{tabular}} & \multicolumn{1}{c}{\begin{tabular}[c]{@{}c@{}}\textbf{LPIPS}\end{tabular}} & \multicolumn{1}{c}{\begin{tabular}[c]{@{}c@{}}\textbf{Enc. Time}\end{tabular}} & \multicolumn{1}{c}{\begin{tabular}[c]{@{}c@{}}\textbf{Dec. Time}\end{tabular}} & \multicolumn{1}{c}{\begin{tabular}[c]{@{}c@{}}\textbf{Mem}\end{tabular}} & \multicolumn{1}{c}{\begin{tabular}[c]{@{}c@{}}\textbf{PSNR}\end{tabular}} & \multicolumn{1}{c}{\begin{tabular}[c]{@{}c@{}}\textbf{SSIM}\end{tabular}} & \multicolumn{1}{c}{\begin{tabular}[c]{@{}c@{}}\textbf{LPIPS}\end{tabular}} & \multicolumn{1}{c}{\begin{tabular}[c]{@{}c@{}}\textbf{Enc. Time}\end{tabular}} & \multicolumn{1}{c}{\begin{tabular}[c]{@{}c@{}}\textbf{Dec. Time}\end{tabular}} & \multicolumn{1}{c}{\begin{tabular}[c]{@{}c@{}}\textbf{Mem}\end{tabular}} \\
\midrule
\rowcolor{gray!15}
Wan2.2\textsuperscript{*} & 0.70\,B & 39.55 & .965 & .017 & \tmo{842.3} & \tmo{2812.4} & 46.1\,GB & 42.69 & .978 & .015 & \tmo{3463.5} & \tmo{13446.6} & 55.9\,GB \\
TAEW2.2\textsuperscript{*} & 0.01\,B & 33.11 & .910 & .062 & \tms{32.5}{25.9} & \tms{42.3}{66.5} & 0.7\,GB & 36.11 & .950 & .052 & \tms{116.2}{29.8} & \tms{154.3}{87.1} & 2.5\,GB \\
TAEW2.2-Super\textsuperscript{*} & 0.02\,B & 34.55 & .926 & .054 & \tms{32.7}{25.8} & \tms{102.8}{27.4} & 0.7\,GB & 37.62 & .958 & .044 & \tms{116.3}{29.8} & \tms{399.4}{33.7} & 2.7\,GB \\
LightTAE\textsuperscript{*} & 0.01\,B & 29.91 & .862 & .106 & \tms{32.5}{25.9} & \tms{42.3}{66.5} & 0.7\,GB & 31.72 & .909 & .104 & \tms{116.6}{29.7} & \tms{154.7}{86.9} & 2.5\,GB \\
Wan2.2\textsuperscript{*} / LightVAE-Lite\textsuperscript{*} & 0.30\,B & 36.23 & .937 & .054 & \tmo{842.3} & \tms{985.5}{2.9} & 19.6\,GB & 39.97 & .967 & .041 & \tmo{3463.5} & \tms{4724.9}{2.8} & 54.1\,GB \\
Wan2.2\textsuperscript{*} / LightVAE-Pro\textsuperscript{*} & 0.50\,B & \textbf{38.82} & \textbf{.960} & \textbf{.029} & \tmo{842.3} & \tms{1743.8}{1.6} & 20.6\,GB & \textbf{42.25} & \textbf{.978} & \textbf{.024} & \tmo{3463.5} & \tms{8351.0}{1.6} & 54.4\,GB \\
Wan2.2\textsuperscript{*} / Turbo-VAED & 0.25\,B & 36.24 & .941 & .046 & \tmo{842.3} & \tms{1734.9}{1.6} & 35.0\,GB & 39.56 & .970 & .039 & \tmo{3463.5} & \tms{8269.8}{1.6} & 54.0\,GB \\
\addlinespace[2pt]
\textbf{Ours}\textsuperscript{*} & 2.47\,B & 38.40 & .957 & .033 & \tms{569.7}{1.5} & \tms{552.4}{5.1} & 7.5\,GB & 42.24 & \textbf{.978} & .025 & \tms{2127.6}{1.6} & \tms{2066.0}{6.5} & 15.7\,GB \\
\midrule
\rowcolor{gray!15}
H3 & 2.60\,B & 35.86 & .939 & .035 & \tmo{3107.5} & \tmo{2152.0} & 9.2\,GB & 38.68 & .962 & .029 & \tmo{11465.1} & \tmo{8026.2} & 15.0\,GB \\
TAEH3 & 0.01\,B & 29.99 & .860 & .104 & \tms{32.7}{95.0} & \tms{44.8}{48.0} & 7.5\,GB & 32.28 & .914 & .092 & \tms{134.4}{85.3} & \tms{176.5}{45.5} & 23.6\,GB \\
LightVAE-Lite & 0.91\,B & 32.54 & .896 & .096 & \tms{1691.8}{1.8} & \tms{730.4}{2.9} & 5.5\,GB & 35.39 & .939 & .070 & \tms{6254.1}{1.8} & \tms{2715.8}{3.0} & 8.2\,GB \\
\begin{tabular}[c]{@{}l@{}}H3 / LynnReal-Omni\end{tabular} & 1.93\,B & 35.94 & .940 & \textbf{.036} & \tmo{3107.5} & \tms{1398.1}{1.5} & 16.0\,GB & 38.77 & .963 & \textbf{.029} & \tmo{11465.1} & \tms{5231.7}{1.5} & 40.3\,GB \\
H3 / Solar-X\textsuperscript{*} & 2.60\,B & 34.46 & .922 & .043 & \tmo{3107.5} & \tms{1332.6}{1.6} & 36.8\,GB & OOM & OOM & OOM & \tmo{OOM} & \tmo{OOM} & OOM \\
\begin{tabular}[c]{@{}l@{}}H3 / StreamRender-H3\textsuperscript{*}\end{tabular} & 1.80\,B & 34.75 & .929 & .043 & \tmo{3107.5} & \tms{1323.9}{1.6} & 16.0\,GB & 37.57 & .957 & .035 & \tmo{11465.1} & \tms{4805.5}{1.7} & 53.2\,GB \\
\addlinespace[2pt]
\textbf{Ours}\textsuperscript{*} & 2.08\,B & \textbf{36.76} & \textbf{.942} & .042 & \tms{514.2}{6.0} & \tms{660.2}{3.3} & 6.5\,GB & 39.05 & \textbf{.966} & .034 & \tms{1856.0}{6.2} & \tms{2445.0}{3.3} & 14.1\,GB \\
\textbf{Ours-Opt}\textsuperscript{*} & 2.08\,B & 36.63 & .941 & .041 & \tms{165.9}{18.7} & \tms{225.9}{9.5} & 7.8\,GB & \textbf{39.07} & \textbf{.966} & .032 & \tms{656.7}{17.5} & \tms{912.2}{8.8} & 15.1\,GB \\
\bottomrule
\end{tabular}}
\end{table*}

\section{Experiments}
\label{sec:exp}

\subsection{Experimental Setup}
\label{sec:setup}

\noindent\textbf{Base VAEs and models.}
To demonstrate the generalizability of our method across distinct latent representations, we distill LiveVAE from two pretrained video VAEs, MiniMax-H3~\citep{minimax2026h3}, denoted H3, and Wan2.2~\citep{wan2025wan}. We call each of them a \emph{base VAE} and its encoder and decoder the \emph{official} encoder and decoder; they are the teachers of \cref{sec:distill}. All student encoders and decoders use 12 Transformer layers and a causal attention window of size $2\times7\times7$. The hidden dimensions of the encoder and the decoder are 2048 and 2560 for H3, and 2560 and 2496 for Wan2.2.

\noindent\textbf{Training data.}
We train on an internal collection of videos and images, filtered for motion intensity and category diversity, with images making up 20\% of the training samples in all three stages. Videos are 4--6\,s clips sampled at 8 to 30 FPS, from which we take 22 frames for H3 and 21 for Wan2.2, at a resolution of 480p or 720p.

\noindent\textbf{Training details.}
All stages use a global batch size of 8 and AdamW with $\beta_1=0.9$, $\beta_2=0.95$, a weight decay of 0.01 and a learning rate of $10^{-4}$, with 1k warmup steps and cosine decay to zero. Stages 1, 2 and 3 take 80k, 200k and 5k steps, and Stage 3 keeps an EMA of the weights with a decay of 0.999. In all stages, the $\ell_1$ and LPIPS terms of \cref{eq:rec} have weights of 1 and $\lambda_{\mathrm{p}}=0.33$; the others are $\lambda_z=1.0$, $\lambda_{\mathrm{KL}}=10^{-3}$ and $\lambda_{\mathrm{adv}}=0.1$.

\noindent\textbf{Baselines.}
We compare with TAEHV~\citep{BoerBohan2025TAEHV}, LightTAE~\citep{lightx2v2025ae}, LightVAE~\citep{gu2026lightvae}, Turbo-VAED~\citep{zou2025turbovaed}, LynnReal-Omni~\citep{mao2026lynnreal}, Solar-X~\citep{chen2026h3streaming} and StreamRender-H3~\citep{streamrenderh3_2026}, using their public weights for the two base VAEs. We run all of them ourselves and evaluate every row of \cref{tab:main} by end to end reconstruction. A decoder only baseline is paired with the official encoder of its base VAE; for ours, we also report the two combinations that replace a single official component (\cref{sec:single_component}). An asterisk (*) after a model name marks a streaming encoder or decoder, \ie a causal one that processes the video chunk by chunk with a bounded cache. Ours-Opt is the infrastructure optimized version of Ours (\cref{sec:infra}); no other row uses inference optimization, and all rows run on the same clips and GPU, in the same precision and software environment.

\noindent\textbf{Evaluation data.}
We evaluate on 200 clips from UltraVideo~\citep{xue2025ultravideo}, of 22 frames for H3 and 21 frames for Wan2.2, each from a different source video and balanced over five motion tiers, at four resolutions: 480p, 720p, 1080p and 2160p. Each resolution is obtained from the source frames in a single step, by bicubic resizing and center cropping; 1080p uses $1920\times1088$, since both base VAEs compress each spatial axis by a factor of 16. \cref{sec:long_clips} repeats the evaluation on longer clips.

\noindent\textbf{Reconstruction metrics.}
We report PSNR (in dB), SSIM and LPIPS between reconstructed and original frames, averaged over the clips at each resolution; higher PSNR and SSIM and lower LPIPS are better. Time is the median (p50) latency of the encoder or the decoder alone on one clip, in milliseconds, measured after warmup on a single NVIDIA H100 GPU rented from RunPod, with batch size~1, and the speedup is the time of the official component divided by that of the method. Mem is the peak GPU memory in GB and Params the parameter count in billions.

\noindent\textbf{Generation metrics.}
For text to video generation, we report the five dimension scores of VBench-2.0~\citep{zheng2025vbench2}, where higher is better. For reference to video generation, we follow CodeVideoBench~\citep{codevideomodel2026} and report its Structure, Style and Quality scores, each from 0 to 100, with Total defined as the mean of the three (\cref{tab:generation}).

\begin{table}[t]
\centering
\fontsize{6.85}{8.2}\selectfont
\setlength{\tabcolsep}{2.4pt}
\caption{Evaluation on the Generation Tasks.}
\vspace{-1em}
\label{tab:generation}
\label{tab:r2v}
\begin{tabularx}{\linewidth}{l *{12}{>{\hsize=0.8\hsize\centering\arraybackslash}X} *{4}{>{\hsize=1.6\hsize\centering\arraybackslash}X}}
\toprule
\multirow{3}{*}[-2.4ex]{\textbf{VAE}} & \multicolumn{12}{c}{\textbf{Text to Video (VBench-2.0~\citep{zheng2025vbench2})}} & \multicolumn{4}{c}{\textbf{Reference to Video (CodeVideoBench~\citep{codevideomodel2026})}} \\
\cmidrule(lr){2-13} \cmidrule(lr){14-17}
 & \multicolumn{6}{c}{\textbf{Wan2.2}} & \multicolumn{6}{c}{\textbf{MiniMax-H3}} & \multicolumn{4}{c}{\textbf{MiniMax-H3}} \\
\cmidrule(lr){2-7} \cmidrule(lr){8-13} \cmidrule(lr){14-17}
 & \textbf{Human} & \textbf{Creat.} & \textbf{Ctrl.} & \textbf{Phys.} & \textbf{Comm.} & \textbf{Total} & \textbf{Human} & \textbf{Creat.} & \textbf{Ctrl.} & \textbf{Phys.} & \textbf{Comm.} & \textbf{Total} & \textbf{Structure} & \textbf{Style} & \textbf{Quality} & \textbf{Total} \\
\midrule
Official & 0.745 & 0.452 & 0.401 & 0.622 & 0.692 & 0.582 & 0.830 & 0.514 & 0.495 & 0.802 & 0.761 & 0.680 & 75.3 & 86.7 & 90.3 & 84.1 \\
Ours & 0.752 & 0.444 & 0.395 & 0.624 & 0.689 & 0.581 & 0.784 & 0.515 & 0.492 & 0.781 & 0.752 & 0.665 & 74.3 & 87.7 & 89.9 & 84.0 \\
\bottomrule
\end{tabularx}
\end{table}

\subsection{Main Results}
\label{sec:main_results}

\subsubsection{Reconstruction Quality}
\label{sec:recon_quality}
In \cref{tab:main}, LiveVAE exceeds the official H3 teacher in PSNR at all four resolutions by 0.07--0.90\,dB, the highest among all methods from 720p upward. Its LPIPS gap to the official H3 VAE narrows from .054 versus .041 at 480p to .034 versus .029 at 2160p and stays close to the streaming decoder only baselines. On both Wan2.2 and H3, Ours outperforms every baseline that replaces both the encoder and the decoder on all metrics, by 3.0--10.5\,dB in PSNR. LightVAE-Pro and Turbo-VAED on Wan2.2 and LynnReal-Omni on H3 reach comparable or slightly higher quality at some resolutions, but they replace only the decoder and keep the official encoder at its original speed.

\subsubsection{Inference Efficiency}
\label{sec:recon_efficiency}
Tiny models such as TAEH3, TAEW2.2 and LightTAE are faster but trail Ours by more than 4\,dB in PSNR. On H3, where official encoding is slower than decoding, Ours accelerates the encoder by $5.2$--$6.2\times$ and the decoder by $3.0$--$3.3\times$, whereas the decoder only LynnReal-Omni, Solar-X and StreamRender-H3 reach only $1.3$--$1.7\times$ on the decoder and keep the official encoder. With the inference optimizations of \cref{sec:infra}, Ours-Opt reaches up to $18.7\times$ and $9.5\times$ within 0.13\,dB PSNR. Ours also halves H3 peak memory at 480p from 8.8 to 4.4\,GB and needs only 14.1\,GB at 2160p, where Solar-X runs out of memory and StreamRender-H3 uses 53.2\,GB. On Wan2.2, Ours speeds up decoding by $4.7$--$6.5\times$ against $1.6$--$1.8\times$ for LightVAE-Pro and Turbo-VAED, speeds up encoding by $1.4$--$1.6\times$, and cuts peak memory at 2160p from 55.9 to 15.7\,GB, where LightVAE-Pro and Turbo-VAED need 54\,GB or more.

\subsubsection{Generation Quality}
\label{sec:gen_results}
\noindent\textbf{Text to video generation.}
We evaluate text to video generation on VBench-2.0~\citep{zheng2025vbench2} in \cref{tab:generation} with frozen few step DiTs: the 3 step FastWan2.2-TI2V-5B~\citep{zhang2025vsa} for Wan2.2 and the 4 step Wuli MiniMax-H3 Turbo LoRA~\citep{wuli2026h3turbo} for H3. Replacing the decoder on Wan2.2 preserves the overall score at 0.581 versus 0.582, with slight gains in Human Fidelity and Physics. On MiniMax-H3, Ours stays within 0.01 of the official decoder on Creativity, Controllability, and Commonsense, with minor drops in Human Fidelity at 0.784 versus 0.830 and Physics at 0.781 versus 0.802.

\noindent\textbf{Reference to video generation.}
Following CodeVideoBench~\citep{codevideomodel2026} with the official MiniMax-H3 weights~\citep{minimax2026h3}, replacing both the H3 encoder and decoder in \cref{tab:generation} yields a nearly identical total of 84.0 versus 84.1, improving Style from 86.7 to 87.7 while preserving Structure and Quality, which confirms that both modules remain well aligned with the frozen DiT latent space.

\subsubsection{Resolution Extrapolation}
\label{sec:recon_resolution}
Trained only at 480p and 720p, LiveVAE extrapolates directly to 1080p and 2160p without finetuning and outperforms the H3 teacher~\citep{minimax2026h3} in PSNR, aided by local spatial windows and DiT aligned 3D RoPE. Because causal window attention scales linearly with token count, scaling from 720p to 2160p increases H3 encoding and decoding latencies by only $7.9\times$ and $8.1\times$.

\subsubsection{Single Component Results}
\label{sec:single_component}
\cref{tab:single} reports the two combinations that replace a single official component with ours, next to the official VAE and our full model, for both base VAEs. A single replaced component keeps its speedup: on H3, our decoder alone is $3.0$--$3.3\times$ and our encoder alone $4.8$--$6.2\times$ faster than the official one, close to the speedups of the full model.

\begin{table*}[!ht]
\centering
\footnotesize
\setlength{\aboverulesep}{0pt}\setlength{\belowrulesep}{0pt}\renewcommand{\arraystretch}{1.02}
\setlength{\tabcolsep}{2.5pt}
\caption{Single component results.}
\vspace{-0.8em}
\label{tab:single}
\resizebox{\textwidth}{!}{%
\begin{tabular}{l c ccc rrr ccc rrr}
\toprule
 &  & \multicolumn{6}{c}{\textbf{480p}} & \multicolumn{6}{c}{\textbf{720p}} \\
\cmidrule(lr){3-8} \cmidrule(lr){9-14}
\textbf{Encoder / Decoder} & \multicolumn{1}{c}{\textbf{Params}} & \multicolumn{1}{c}{\textbf{PSNR}} & \multicolumn{1}{c}{\textbf{SSIM}} & \multicolumn{1}{c}{\textbf{LPIPS}} & \multicolumn{1}{c}{\textbf{Enc. Time}} & \multicolumn{1}{c}{\textbf{Dec. Time}} & \multicolumn{1}{c}{\textbf{Mem}} & \multicolumn{1}{c}{\textbf{PSNR}} & \multicolumn{1}{c}{\textbf{SSIM}} & \multicolumn{1}{c}{\textbf{LPIPS}} & \multicolumn{1}{c}{\textbf{Enc. Time}} & \multicolumn{1}{c}{\textbf{Dec. Time}} & \multicolumn{1}{c}{\textbf{Mem}} \\
\midrule
\rowcolor{gray!15}
Wan2.2\textsuperscript{*} & 0.70\,B & 36.08 & .945 & .019 & \tmo{175.0} & \tmo{560.3} & 11.0\,GB & 37.66 & .954 & .018 & \tmo{382.8} & \tmo{1264.2} & 21.8\,GB \\
Wan2.2\textsuperscript{*} / \textbf{Ours}\textsuperscript{*} & 1.35\,B & 34.11 & .922 & .043 & \tmo{175.0} & \tms{121.2}{4.6} & 5.6\,GB & 35.86 & .938 & .039 & \tmo{382.8} & \tms{254.7}{5.0} & 9.2\,GB \\
\textbf{Ours}\textsuperscript{*} / Wan2.2\textsuperscript{*} & 1.82\,B & \textbf{35.16} & \textbf{.937} & \textbf{.022} & \tms{127.2}{1.4} & \tmo{560.3} & 12.1\,GB & \textbf{36.85} & \textbf{.949} & \textbf{.021} & \tms{262.1}{1.5} & \tmo{1264.2} & 22.1\,GB \\
\textbf{Ours}\textsuperscript{*} & 2.47\,B & 34.46 & .927 & .042 & \tms{124.9}{1.4} & \tms{119.5}{4.7} & 5.2\,GB & 36.24 & .941 & .038 & \tms{261.5}{1.5} & \tms{254.9}{5.0} & 5.9\,GB \\
\midrule
\rowcolor{gray!15}
H3 & 2.60\,B & 32.99 & .909 & .041 & \tmo{623.0} & \tmo{435.3} & 8.8\,GB & 34.31 & .923 & .038 & \tmo{1453.5} & \tmo{1010.6} & 9.0\,GB \\
H3 / \textbf{Ours}\textsuperscript{*} & 1.45\,B & 32.35 & .896 & .058 & \tmo{623.0} & \tms{146.2}{3.0} & 6.7\,GB & 33.94 & .915 & .052 & \tmo{1453.5} & \tms{304.4}{3.3} & 6.8\,GB \\
\textbf{Ours}\textsuperscript{*} / H3 & 3.24\,B & 32.08 & .895 & \textbf{.049} & \tms{130.3}{4.8} & \tmo{435.3} & 6.7\,GB & 33.46 & .912 & \textbf{.046} & \tms{235.6}{6.2} & \tmo{1010.6} & 7.4\,GB \\
\textbf{Ours}\textsuperscript{*} & 2.08\,B & \textbf{33.06} & \textbf{.904} & .054 & \tms{119.5}{5.2} & \tms{146.0}{3.0} & 4.4\,GB & \textbf{34.74} & \textbf{.922} & .049 & \tms{234.3}{6.2} & \tms{303.4}{3.3} & 5.0\,GB \\
\specialrule{\heavyrulewidth}{0pt}{0pt}
 &  & \multicolumn{6}{c}{\textbf{1080p}} & \multicolumn{6}{c}{\textbf{2160p}} \\
\cmidrule(lr){3-8} \cmidrule(lr){9-14}
\textbf{Encoder / Decoder} & \multicolumn{1}{c}{\textbf{Params}} & \multicolumn{1}{c}{\textbf{PSNR}} & \multicolumn{1}{c}{\textbf{SSIM}} & \multicolumn{1}{c}{\textbf{LPIPS}} & \multicolumn{1}{c}{\textbf{Enc. Time}} & \multicolumn{1}{c}{\textbf{Dec. Time}} & \multicolumn{1}{c}{\textbf{Mem}} & \multicolumn{1}{c}{\textbf{PSNR}} & \multicolumn{1}{c}{\textbf{SSIM}} & \multicolumn{1}{c}{\textbf{LPIPS}} & \multicolumn{1}{c}{\textbf{Enc. Time}} & \multicolumn{1}{c}{\textbf{Dec. Time}} & \multicolumn{1}{c}{\textbf{Mem}} \\
\midrule
\rowcolor{gray!15}
Wan2.2\textsuperscript{*} & 0.70\,B & 39.55 & .965 & .017 & \tmo{842.3} & \tmo{2812.4} & 46.1\,GB & 42.69 & .978 & .015 & \tmo{3463.5} & \tmo{13446.6} & 55.9\,GB \\
Wan2.2\textsuperscript{*} / \textbf{Ours}\textsuperscript{*} & 1.35\,B & 37.98 & .954 & .033 & \tmo{842.3} & \tms{557.2}{5.0} & 17.2\,GB & 41.74 & .976 & .025 & \tmo{3463.5} & \tms{2075.2}{6.5} & 59.9\,GB \\
\textbf{Ours}\textsuperscript{*} / Wan2.2\textsuperscript{*} & 1.82\,B & \textbf{38.86} & \textbf{.962} & \textbf{.019} & \tms{574.6}{1.5} & \tmo{2812.4} & 44.3\,GB & 42.23 & \textbf{.978} & \textbf{.016} & \tms{2128.7}{1.6} & \tmo{13446.6} & 53.6\,GB \\
\textbf{Ours}\textsuperscript{*} & 2.47\,B & 38.40 & .957 & .033 & \tms{569.7}{1.5} & \tms{552.4}{5.1} & 7.5\,GB & \textbf{42.24} & \textbf{.978} & .025 & \tms{2127.6}{1.6} & \tms{2066.0}{6.5} & 15.7\,GB \\
\midrule
\rowcolor{gray!15}
H3 & 2.60\,B & 35.86 & .939 & .035 & \tmo{3107.5} & \tmo{2152.0} & 9.2\,GB & 38.68 & .962 & .029 & \tmo{11465.1} & \tmo{8026.2} & 15.0\,GB \\
H3 / \textbf{Ours}\textsuperscript{*} & 1.45\,B & 35.83 & .935 & .044 & \tmo{3107.5} & \tms{661.3}{3.3} & 7.1\,GB & \textbf{39.34} & .965 & \textbf{.032} & \tmo{11465.1} & \tms{2452.5}{3.3} & 9.8\,GB \\
\textbf{Ours}\textsuperscript{*} / H3 & 3.24\,B & 34.97 & .931 & \textbf{.042} & \tms{514.1}{6.0} & \tmo{2152.0} & 8.8\,GB & 36.56 & .952 & .042 & \tms{1859.9}{6.2} & \tmo{8026.2} & 16.3\,GB \\
\textbf{Ours}\textsuperscript{*} & 2.08\,B & \textbf{36.76} & \textbf{.942} & \textbf{.042} & \tms{514.2}{6.0} & \tms{660.2}{3.3} & 6.5\,GB & 39.05 & \textbf{.966} & .034 & \tms{1856.0}{6.2} & \tms{2445.0}{3.3} & 14.1\,GB \\
\bottomrule
\end{tabular}}
\end{table*}

\subsubsection{Longer Clips}
\label{sec:long_clips}
\cref{tab:main} evaluates short clips of 22 frames for H3 and 21 frames for Wan2.2. \cref{tab:long} repeats the same comparison on longer clips: all H3 rows are evaluated on clips of 90 frames and all Wan2.2 rows on clips of 89 frames, about four times the length of the short clips. All other settings follow \cref{sec:setup}. On Wan2.2, our encoder is $1.5$--$1.7\times$ and our decoder $5.0$--$7.1\times$ faster than the official ones ($1.4$--$1.6\times$ and $4.7$--$6.5\times$ on short clips), and peak memory falls from 11.0 to 5.2\,GB at 480p and from 60.9 to 15.7\,GB at 2160p, at a PSNR 1.4 to 2.2\,dB below the official VAE. On H3, \cref{tab:long} lists Ours and Ours-Opt separately. Ours follows the protocol of the baselines: its encoder is $5.4$--$6.4\times$ and its decoder $3.8$--$4.2\times$ faster than the official ones ($5.2$--$6.2\times$ and $3.0$--$3.3\times$ on short clips), and its PSNR is 0.3 to 1.3\,dB above the official VAE up to 1080p and within 0.05\,dB at 2160p, while LPIPS remains higher. Ours-Opt adds the inference optimizations of \cref{sec:infra}, raising the speedup to $8.0$--$18.9\times$ for the encoder and $6.9$--$11.9\times$ for the decoder at a PSNR within 0.16\,dB of Ours. The closest baseline in quality is the decoder only LynnReal-Omni, which reaches a $1.5\times$ decoder speedup and keeps the official encoder. At 2160p, Solar-X runs out of memory on H3 and Turbo-VAED already does so from 1080p on Wan2.2, whereas Ours needs 14.4 and 15.7\,GB.

\begin{table*}[!ht]
\centering
\footnotesize
\setlength{\aboverulesep}{0pt}\setlength{\belowrulesep}{0pt}\renewcommand{\arraystretch}{1.02}
\setlength{\tabcolsep}{2.5pt}
\caption{Reconstruction on longer clips of 90 frames (H3) and 89 frames (Wan2.2).}
\vspace{-0.8em}
\label{tab:long}
\resizebox{\textwidth}{!}{%
\begin{tabular}{l c ccc rrr ccc rrr}
\toprule
 &  & \multicolumn{6}{c}{\textbf{480p}} & \multicolumn{6}{c}{\textbf{720p}} \\
\cmidrule(lr){3-8} \cmidrule(lr){9-14}
\textbf{Encoder / Decoder} & \multicolumn{1}{c}{\textbf{Params}} & \multicolumn{1}{c}{\textbf{PSNR}} & \multicolumn{1}{c}{\textbf{SSIM}} & \multicolumn{1}{c}{\textbf{LPIPS}} & \multicolumn{1}{c}{\textbf{Enc. Time}} & \multicolumn{1}{c}{\textbf{Dec. Time}} & \multicolumn{1}{c}{\textbf{Mem}} & \multicolumn{1}{c}{\textbf{PSNR}} & \multicolumn{1}{c}{\textbf{SSIM}} & \multicolumn{1}{c}{\textbf{LPIPS}} & \multicolumn{1}{c}{\textbf{Enc. Time}} & \multicolumn{1}{c}{\textbf{Dec. Time}} & \multicolumn{1}{c}{\textbf{Mem}} \\
\midrule
\rowcolor{gray!15}
Wan2.2\textsuperscript{*} & 0.70\,B & 35.22 & .931 & .023 & \tmo{724.5} & \tmo{2354.4} & 11.0\,GB & 36.75 & .945 & .023 & \tmo{1599.5} & \tmo{5327.8} & 21.8\,GB \\
TAEW2.2\textsuperscript{*} & 0.01\,B & 29.49 & .837 & .089 & \tms{47.1}{15.4} & \tms{59.3}{39.7} & 0.2\,GB & 30.85 & .867 & .083 & \tms{67.0}{23.9} & \tms{85.8}{62.1} & 0.3\,GB \\
TAEW2.2-Super\textsuperscript{*} & 0.02\,B & 30.76 & .862 & .080 & \tms{41.9}{17.3} & \tms{82.1}{28.7} & 0.2\,GB & 32.20 & .888 & .073 & \tms{66.6}{24.0} & \tms{177.8}{30.0} & 0.3\,GB \\
LightTAE\textsuperscript{*} & 0.01\,B & 27.63 & .794 & .118 & \tms{53.1}{13.6} & \tms{62.0}{38.0} & 0.2\,GB & 28.51 & .822 & .117 & \tms{67.6}{23.7} & \tms{86.4}{61.7} & 0.3\,GB \\
Wan2.2\textsuperscript{*} / LightVAE-Lite\textsuperscript{*} & 0.30\,B & 32.26 & .884 & .083 & \tmo{724.5} & \tms{816.7}{2.9} & 4.4\,GB & 33.92 & .907 & .073 & \tmo{1599.5} & \tms{1836.6}{2.9} & 9.1\,GB \\
Wan2.2\textsuperscript{*} / LightVAE-Pro\textsuperscript{*} & 0.50\,B & \textbf{34.64} & \textbf{.921} & \textbf{.040} & \tmo{724.5} & \tms{1492.7}{1.6} & 4.9\,GB & \textbf{36.23} & \textbf{.937} & \textbf{.038} & \tmo{1599.5} & \tms{3308.4}{1.6} & 9.8\,GB \\
Wan2.2\textsuperscript{*} / Turbo-VAED & 0.25\,B & 30.85 & .886 & .070 & \tmo{724.5} & \tms{1354.1}{1.7} & 31.6\,GB & 31.94 & .907 & .067 & \tmo{1599.5} & \tms{7075.0}{0.8} & 64.3\,GB \\
\textbf{Ours}\textsuperscript{*} & 2.47\,B & 32.98 & .896 & .053 & \tms{485.4}{1.5} & \tms{466.9}{5.0} & 5.2\,GB & 34.68 & .918 & .048 & \tms{1020.0}{1.6} & \tms{995.0}{5.4} & 5.9\,GB \\
\midrule
\rowcolor{gray!15}
H3 & 2.60\,B & 34.09 & .915 & .042 & \tmo{2482.7} & \tmo{2151.3} & 9.0\,GB & 35.39 & .929 & .040 & \tmo{5827.0} & \tmo{5007.6} & 9.3\,GB \\
TAEH3 & 0.01\,B & 28.48 & .807 & .123 & \tms{27.0}{92.0} & \tms{33.1}{65.0} & 5.5\,GB & 29.69 & .839 & .116 & \tms{57.8}{100.8} & \tms{79.8}{62.8} & 12.7\,GB \\
LightVAE-Lite & 0.91\,B & 30.63 & .856 & .123 & \tms{1349.8}{1.8} & \tms{728.3}{3.0} & 5.3\,GB & 32.14 & .883 & .108 & \tms{3160.9}{1.8} & \tms{1698.2}{2.9} & 5.6\,GB \\
H3 / LynnReal-Omni & 1.93\,B & 34.26 & \textbf{.918} & \textbf{.045} & \tmo{2482.7} & \tms{1405.8}{1.5} & 11.0\,GB & 35.48 & \textbf{.931} & \textbf{.042} & \tmo{5827.0} & \tms{3292.4}{1.5} & 11.7\,GB \\
H3 / Solar-X\textsuperscript{*} & 2.60\,B & 33.01 & .896 & .048 & \tmo{2482.7} & \tms{1245.2}{1.7} & 11.6\,GB & 34.29 & .913 & .046 & \tmo{5827.0} & \tms{2904.8}{1.7} & 20.6\,GB \\
H3 / StreamRender-H3\textsuperscript{*} & 1.80\,B & 32.92 & .900 & .050 & \tmo{2482.7} & \tms{1257.2}{1.7} & 7.5\,GB & 34.13 & .916 & .049 & \tmo{5827.0} & \tms{2446.9}{2.0} & 9.2\,GB \\
\textbf{Ours}\textsuperscript{*} & 2.08\,B & \textbf{34.41} & .910 & .052 & \tms{457.0}{5.4} & \tms{565.2}{3.8} & 4.4\,GB & \textbf{36.19} & .928 & .048 & \tms{915.7}{6.4} & \tms{1187.4}{4.2} & 5.1\,GB \\
\textbf{Ours-Opt}\textsuperscript{*} & 2.08\,B & \textbf{34.41} & .910 & .052 & \tms{310.4}{8.0} & \tms{313.0}{6.9} & 4.4\,GB & \textbf{36.19} & .928 & .048 & \tms{370.2}{15.7} & \tms{511.2}{9.8} & 5.0\,GB \\
\specialrule{\heavyrulewidth}{0pt}{0pt}
 &  & \multicolumn{6}{c}{\textbf{1080p}} & \multicolumn{6}{c}{\textbf{2160p}} \\
\cmidrule(lr){3-8} \cmidrule(lr){9-14}
\textbf{Encoder / Decoder} & \multicolumn{1}{c}{\textbf{Params}} & \multicolumn{1}{c}{\textbf{PSNR}} & \multicolumn{1}{c}{\textbf{SSIM}} & \multicolumn{1}{c}{\textbf{LPIPS}} & \multicolumn{1}{c}{\textbf{Enc. Time}} & \multicolumn{1}{c}{\textbf{Dec. Time}} & \multicolumn{1}{c}{\textbf{Mem}} & \multicolumn{1}{c}{\textbf{PSNR}} & \multicolumn{1}{c}{\textbf{SSIM}} & \multicolumn{1}{c}{\textbf{LPIPS}} & \multicolumn{1}{c}{\textbf{Enc. Time}} & \multicolumn{1}{c}{\textbf{Dec. Time}} & \multicolumn{1}{c}{\textbf{Mem}} \\
\midrule
\rowcolor{gray!15}
Wan2.2\textsuperscript{*} & 0.70\,B & 38.60 & .959 & .020 & \tmo{3549.4} & \tmo{12037.7} & 46.1\,GB & 41.83 & .976 & .017 & \tmo{14567.9} & \tmo{57208.7} & 60.9\,GB \\
TAEW2.2\textsuperscript{*} & 0.01\,B & 32.42 & .899 & .074 & \tms{123.0}{28.9} & \tms{161.1}{74.7} & 0.7\,GB & 35.26 & .944 & .061 & \tms{442.3}{32.9} & \tms{590.9}{96.8} & 2.5\,GB \\
TAEW2.2-Super\textsuperscript{*} & 0.02\,B & 33.84 & .915 & .064 & \tms{121.7}{29.2} & \tms{393.1}{30.6} & 0.7\,GB & 36.82 & .954 & .051 & \tms{443.8}{32.8} & \tms{1532.1}{37.3} & 2.7\,GB \\
LightTAE\textsuperscript{*} & 0.01\,B & 29.54 & .854 & .116 & \tms{123.3}{28.8} & \tms{161.3}{74.6} & 0.7\,GB & 31.33 & .904 & .116 & \tms{444.3}{32.8} & \tms{592.9}{96.5} & 2.5\,GB \\
Wan2.2\textsuperscript{*} / LightVAE-Lite\textsuperscript{*} & 0.30\,B & 35.80 & .932 & .061 & \tmo{3549.4} & \tms{4133.1}{2.9} & 19.6\,GB & 39.28 & .965 & .044 & \tmo{14567.9} & \tms{19941.8}{2.9} & 54.1\,GB \\
Wan2.2\textsuperscript{*} / LightVAE-Pro\textsuperscript{*} & 0.50\,B & \textbf{38.11} & \textbf{.955} & \textbf{.033} & \tmo{3549.4} & \tms{7338.1}{1.6} & 20.6\,GB & \textbf{41.52} & \textbf{.976} & \textbf{.025} & \tmo{14567.9} & \tms{35357.0}{1.6} & 54.5\,GB \\
Wan2.2\textsuperscript{*} / Turbo-VAED & 0.25\,B & OOM & OOM & OOM & \tmo{OOM} & \tmo{OOM} & OOM & OOM & OOM & OOM & \tmo{OOM} & \tmo{OOM} & OOM \\
\textbf{Ours}\textsuperscript{*} & 2.47\,B & 36.66 & .942 & .041 & \tms{2243.6}{1.6} & \tms{2175.3}{5.5} & 7.5\,GB & 40.41 & .973 & .029 & \tms{8332.3}{1.7} & \tms{8093.7}{7.1} & 15.7\,GB \\
\midrule
\rowcolor{gray!15}
H3 & 2.60\,B & 36.76 & .944 & .037 & \tmo{12456.8} & \tmo{10679.7} & 10.0\,GB & 39.10 & .965 & .032 & \tmo{45807.1} & \tmo{39544.8} & 20.1\,GB \\
TAEH3 & 0.01\,B & 31.06 & .875 & .104 & \tms{124.0}{100.5} & \tms{178.7}{59.8} & 24.5\,GB & 32.96 & .922 & .094 & \tms{522.6}{87.7} & \tms{1553.0}{25.5} & 73.6\,GB \\
LightVAE-Lite & 0.91\,B & 33.71 & .909 & .091 & \tms{6772.8}{1.8} & \tms{3650.1}{2.9} & 8.5\,GB & 36.19 & .947 & .070 & \tms{24848.3}{1.8} & \tms{13375.4}{3.0} & 28.5\,GB \\
H3 / LynnReal-Omni & 1.93\,B & 36.82 & .945 & \textbf{.038} & \tmo{12456.8} & \tms{6915.6}{1.5} & 17.3\,GB & \textbf{39.20} & .966 & \textbf{.032} & \tmo{45807.1} & \tms{25644.2}{1.5} & 45.3\,GB \\
H3 / Solar-X\textsuperscript{*} & 2.60\,B & 35.65 & .932 & .041 & \tmo{12456.8} & \tms{6018.0}{1.8} & 36.8\,GB & OOM & OOM & OOM & \tmo{OOM} & \tmo{OOM} & OOM \\
H3 / StreamRender-H3\textsuperscript{*} & 1.80\,B & 35.62 & .935 & .044 & \tmo{12456.8} & \tms{5061.6}{2.1} & 16.2\,GB & 37.98 & .961 & .038 & \tmo{45807.1} & \tms{18441.1}{2.1} & 53.8\,GB \\
\textbf{Ours}\textsuperscript{*} & 2.08\,B & \textbf{38.08} & \textbf{.946} & .042 & \tms{2018.7}{6.2} & \tms{2593.7}{4.1} & 6.6\,GB & 39.05 & \textbf{.967} & .039 & \tms{7298.7}{6.3} & \tms{9596.1}{4.1} & 14.4\,GB \\
\textbf{Ours-Opt}\textsuperscript{*} & 2.08\,B & 37.92 & \textbf{.946} & .041 & \tms{658.8}{18.9} & \tms{898.9}{11.9} & 7.8\,GB & 38.94 & \textbf{.967} & .037 & \tms{2538.1}{18.0} & \tms{3525.7}{11.2} & 15.3\,GB \\
\bottomrule
\end{tabular}}
\end{table*}

\subsection{Ablation Studies}
\label{sec:ablation}
\begin{figure}[t]
\centering
\begin{minipage}[t]{0.403\linewidth}\centering
\includegraphics[width=\linewidth]{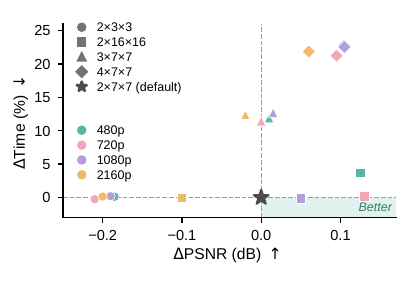}
\caption{\textbf{Attention window tradeoff.} Larger spatial windows add little PSNR, and longer temporal ones cost over 11\% more time.}
  \label{fig:window_tradeoff}
\end{minipage}\hfill
\begin{minipage}[t]{0.577\linewidth}\centering
\includegraphics[width=\linewidth]{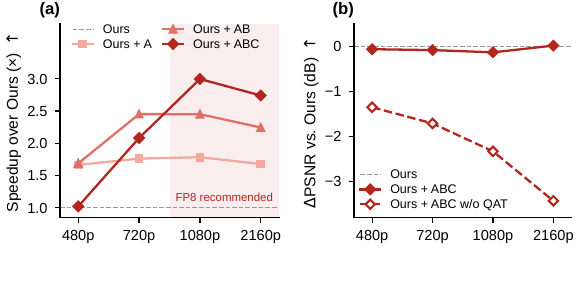}
\caption{\textbf{Cumulative inference optimizations.} (a) End to end speedup over Ours. (b) PSNR change relative to Ours. A: attention operator fusion. B: region wise compilation. C: FFN FP8 with QAT.}
\label{fig:infra}
\end{minipage}
\end{figure}

All ablations are conducted on MiniMax-H3, examining our causal window design (\cref{sec:na}, \cref{fig:window_tradeoff}), three stage distillation (\cref{sec:distill}, \cref{tab:data,tab:joint,tab:gan}), and inference infrastructure (\cref{sec:infra}, \cref{fig:infra}). Unless stated otherwise, they are evaluated on the 200 UltraVideo clips of \cref{sec:setup} at all four resolutions, and gray rows mark the setting selected for our final model.

\subsubsection{Ablation on Causal Window Attention Design}
\label{sec:abl_window_design}
We vary spatial and temporal window sizes to validate our effective receptive field findings (\cref{fig:motivation}). Each component is trained separately with a shortened schedule, 60k steps for encoders and 120k steps for decoders, and evaluated with its official counterpart, with compilation off. \cref{tab:window,tab:causal} in \cref{sec:app_window} give the per component numbers.

\noindent\textbf{Attention window tradeoff.}
\label{sec:abl_window_tradeoff}
\label{sec:abl_window}
\label{sec:abl_causal}
\cref{fig:window_tradeoff} plots each candidate window against our default $2\times7\times7$, averaged over the encoder and decoder; we cite per component numbers below. Shrinking the spatial window to $3\times3$ saves no latency and costs up to 0.08\,dB on the decoder and 0.33--0.36\,dB on the encoder at every resolution. We choose $7\times7$ over $16\times16$ for three reasons. First, $16\times16$ does not extrapolate: its encoder leads $7\times7$ by 0.12\,dB at the 480p and 720p training resolutions but trails by 0.06\,dB at 1080p and 0.34\,dB at 2160p. Second, $7\times7$ shows no such drop, so one model scales directly from low to high resolution. Third, the gain of $16\times16$ is limited, 0.13--0.16\,dB on the decoder with LPIPS lower by only 0.001. Temporally, extending $T$ from 2 to 3 or 4 latent frames increases latency by 11--13\% or 21--23\% while shifting PSNR by at most 0.22\,dB.

\noindent\textbf{Frame wise 2D decomposition.}
\label{sec:abl_decomp}
We compare joint 3D window attention against our mathematically equivalent frame wise 2D decomposition (\cref{eq:per_frame,eq:merge}) using identical weights, so reconstruction quality is unchanged. In \cref{tab:decomp}, the joint 3D kernel is 27--28\% slower and uses 18--21\% more memory at 720p; across all four resolutions, it is 21--29\% slower and consumes 9--53\% more memory, validating our decomposed formulation.

\begin{table*}[!ht]
\centering
\footnotesize
\setlength{\aboverulesep}{0pt}\setlength{\belowrulesep}{0pt}\renewcommand{\arraystretch}{1.02}
\setlength{\tabcolsep}{4pt}
\caption{Attention decomposition.}
\vspace{-0.8em}
\label{tab:decomp}
\begin{tabularx}{\textwidth}{l l YY YY YY YY}
\toprule
\multirow{2}{*}{Module} & \multirow{2}{*}{Attention} & \multicolumn{2}{c}{480p} & \multicolumn{2}{c}{720p} & \multicolumn{2}{c}{1080p} & \multicolumn{2}{c}{2160p} \\
\cmidrule(lr){3-4} \cmidrule(lr){5-6} \cmidrule(lr){7-8} \cmidrule(lr){9-10}
& & Time & Mem & Time & Mem & Time & Mem & Time & Mem \\
\midrule
& Joint 3D & 182.2 & 2.96 & 384.0 & 3.72 & 796.9 & 5.27 & 3106.5 & 14.20 \\
\rowcolor{gray!15}
\cellcolor{white}\multirow{-2}{*}{Decoder} & \textbf{Decomposed} & \textbf{146.0} & \textbf{2.71} & \textbf{303.4} & \textbf{3.14} & \textbf{660.2} & \textbf{4.13} & \textbf{2445.0} & \textbf{9.36} \\
\midrule
& Joint 3D & 149.2 & 2.00 & 299.5 & 2.62 & 623.0 & 3.90 & 2387.8 & 11.22 \\
\rowcolor{gray!15}
\cellcolor{white}\multirow{-2}{*}{Encoder} & \textbf{Decomposed} & \textbf{119.5} & \textbf{1.80} & \textbf{234.3} & \textbf{2.16} & \textbf{514.2} & \textbf{2.98} & \textbf{1856.0} & \textbf{7.34} \\
\bottomrule
\end{tabularx}
\end{table*}

\noindent\textbf{RoPE selection.}
\label{sec:abl_rope}
All CWA blocks use the 3D RoPE of the DiT of the base model (\cref{sec:encdec}). \cref{tab:rope} compares this choice on H3 with the 3D RoPE of Wan2.2, evaluated end to end. The H3 DiT RoPE row is the Ours row of \cref{tab:main}. The difference is small: with the Wan2.2 RoPE, PSNR is lower by 0.11 to 0.23\,dB and LPIPS higher by 0.002 to 0.005, while SSIM changes by less than 0.001. The PSNR gap tends to grow with resolution, from 0.11\,dB at 480p to 0.23\,dB at 2160p. Both variants work, and following the DiT appears to be slightly better, so we keep it as the default for both base VAEs.

\begin{table*}[!ht]
\centering
\footnotesize
\setlength{\aboverulesep}{0pt}\setlength{\belowrulesep}{0pt}\renewcommand{\arraystretch}{1.02}
\setlength{\tabcolsep}{4pt}
\caption{RoPE selection.}
\vspace{-0.8em}
\label{tab:rope}
\begin{tabularx}{\textwidth}{l YYY YYY YYY YYY}
\toprule
\multirow{2}{*}{RoPE} & \multicolumn{3}{c}{480p} & \multicolumn{3}{c}{720p} & \multicolumn{3}{c}{1080p} & \multicolumn{3}{c}{2160p} \\
\cmidrule(lr){2-4} \cmidrule(lr){5-7} \cmidrule(lr){8-10} \cmidrule(lr){11-13}
 & PSNR & SSIM & LPIPS & PSNR & SSIM & LPIPS & PSNR & SSIM & LPIPS & PSNR & SSIM & LPIPS \\
\midrule
\rowcolor{gray!15}
\textbf{H3 DiT RoPE} & 33.06 & 0.904 & 0.054 & 34.74 & 0.922 & 0.049 & 36.76 & 0.942 & 0.042 & 39.05 & 0.966 & 0.034 \\
Wan2.2 RoPE & 32.95 & 0.903 & 0.059 & 34.61 & 0.921 & 0.053 & 36.60 & 0.941 & 0.045 & 38.82 & 0.966 & 0.035 \\
\bottomrule
\end{tabularx}
\end{table*}

\subsubsection{Ablation on Progressive VAE Distillation}
\label{sec:abl_training}
These ablations evaluate the key components of our three stage distillation framework in \cref{sec:distill}: the image and video mixing ratio, joint finetuning in Stage~3, and the choice of adversarial discriminator.

\noindent\textbf{Image and video mix.}
\label{sec:abl_data}
\cref{tab:data} varies the fraction of images in the training data. At 720p, mixing in 20\% images lowers LPIPS from 0.054 to 0.049 at a 0.08\,dB PSNR cost, as high resolution still images provide sharper spatial supervision. A 50\% ratio brings no further gain, so we adopt 20\%.

\begin{table*}[!ht]
\centering
\footnotesize
\setlength{\aboverulesep}{0pt}\setlength{\belowrulesep}{0pt}\renewcommand{\arraystretch}{1.02}
\setlength{\tabcolsep}{4pt}
\caption{Image and video mix.}
\vspace{-0.8em}
\label{tab:data}
\begin{tabularx}{\textwidth}{c YYY YYY YYY YYY}
\toprule
\multirow{2}{*}{Images} & \multicolumn{3}{c}{480p} & \multicolumn{3}{c}{720p} & \multicolumn{3}{c}{1080p} & \multicolumn{3}{c}{2160p} \\
\cmidrule(lr){2-4} \cmidrule(lr){5-7} \cmidrule(lr){8-10} \cmidrule(lr){11-13}
 & PSNR & SSIM & LPIPS & PSNR & SSIM & LPIPS & PSNR & SSIM & LPIPS & PSNR & SSIM & LPIPS \\
\midrule
0\% & 33.14 & 0.905 & 0.060 & 34.82 & 0.923 & 0.054 & 36.83 & 0.942 & 0.046 & 39.11 & 0.966 & 0.037 \\
\rowcolor{gray!15}
\textbf{20\%} & 33.06 & 0.904 & 0.054 & 34.74 & 0.922 & 0.049 & 36.76 & 0.942 & 0.042 & 39.05 & 0.966 & 0.034 \\
50\% & 33.05 & 0.904 & 0.054 & 34.73 & 0.922 & 0.049 & 36.75 & 0.942 & 0.042 & 39.02 & 0.966 & 0.034 \\
\bottomrule
\end{tabularx}
\end{table*}

\noindent\textbf{Joint finetuning.}
\label{sec:abl_joint}
\cref{tab:joint} ablates joint finetuning on end to end reconstruction. Without it, the Stage~1 encoder is paired directly with the Stage~2 decoder, which was trained only on teacher latents and has never seen student latents. Joint finetuning with \cref{eq:joint} lets the two modules adapt to each other, raising PSNR at 720p by 1.08\,dB from 33.66 to 34.74\,dB and lowering LPIPS from 0.062 to 0.049. The gain holds at all four resolutions: PSNR rises by 0.32 to 1.29\,dB and LPIPS falls by 11 to 21\%. Before joint finetuning, the PSNR of our model is below that of the official VAE up to 1080p; after it, it is 0.07 to 0.90\,dB above at all four resolutions, while its LPIPS remains higher. The cost is a minor drop when either module is paired with its official counterpart (\cref{tab:single}).

\begin{table*}[!ht]
\centering
\footnotesize
\setlength{\aboverulesep}{0pt}\setlength{\belowrulesep}{0pt}\renewcommand{\arraystretch}{1.02}
\setlength{\tabcolsep}{4pt}
\caption{Joint finetuning.}
\vspace{-0.8em}
\label{tab:joint}
\begin{tabularx}{\textwidth}{l YYY YYY YYY YYY}
\toprule
\multirow{2}{*}{Training} & \multicolumn{3}{c}{480p} & \multicolumn{3}{c}{720p} & \multicolumn{3}{c}{1080p} & \multicolumn{3}{c}{2160p} \\
\cmidrule(lr){2-4} \cmidrule(lr){5-7} \cmidrule(lr){8-10} \cmidrule(lr){11-13}
 & PSNR & SSIM & LPIPS & PSNR & SSIM & LPIPS & PSNR & SSIM & LPIPS & PSNR & SSIM & LPIPS \\
\midrule
Official H3 & 32.99 & 0.909 & 0.041 & 34.31 & 0.923 & 0.038 & 35.86 & 0.939 & 0.035 & 38.68 & 0.962 & 0.029 \\
\midrule
w/o Joint Finetuning & 32.21 & 0.893 & 0.069 & 33.66 & 0.912 & 0.062 & 35.47 & 0.933 & 0.053 & 38.73 & 0.964 & 0.038 \\
\rowcolor{gray!15}
\textbf{w/ Joint Finetuning} & 33.06 & 0.904 & 0.054 & 34.74 & 0.922 & 0.049 & 36.76 & 0.942 & 0.042 & 39.05 & 0.966 & 0.034 \\
\bottomrule
\end{tabularx}
\end{table*}

\noindent\textbf{Adversarial loss.}
\label{sec:abl_gan}
\cref{tab:gan} compares discriminators for the adversarial loss in joint finetuning, whose purpose is to sharpen fine textures, measured by LPIPS. PatchGAN discriminators~\citep{isola2017pix2pix} trained from scratch hardly help and can even hurt: at 720p, LPIPS stays at 0.057 or rises to 0.063. They have not learned representations that capture the fine structural detail of ViT outputs, so their feedback does not target the blur. The pretrained Real-ESRGAN U-Net discriminator~\citep{wang2021realesrgan} already separates high frequency detail, so it corrects blurry results and missing detail, lowering LPIPS from 0.057 to 0.049 while trading off a negligible 0.07\,dB in PSNR at 34.74\,dB against 34.81\,dB as sharper synthesized textures deviate slightly from pixelwise averages.

\begin{table*}[!ht]
\centering
\footnotesize
\setlength{\aboverulesep}{0pt}\setlength{\belowrulesep}{0pt}\renewcommand{\arraystretch}{1.02}
\setlength{\tabcolsep}{4pt}
\caption{Adversarial loss.}
\vspace{-0.8em}
\label{tab:gan}
\begin{tabularx}{\textwidth}{l YYY YYY YYY YYY}
\toprule
\multirow{2}{*}{Discriminator} & \multicolumn{3}{c}{480p} & \multicolumn{3}{c}{720p} & \multicolumn{3}{c}{1080p} & \multicolumn{3}{c}{2160p} \\
\cmidrule(lr){2-4} \cmidrule(lr){5-7} \cmidrule(lr){8-10} \cmidrule(lr){11-13}
 & PSNR & SSIM & LPIPS & PSNR & SSIM & LPIPS & PSNR & SSIM & LPIPS & PSNR & SSIM & LPIPS \\
\midrule
None & 33.13 & 0.905 & 0.063 & 34.81 & 0.923 & 0.057 & 36.82 & 0.943 & 0.049 & 39.11 & 0.967 & 0.040 \\
2D PatchGAN & 33.12 & 0.908 & 0.063 & 34.78 & 0.921 & 0.057 & 36.80 & 0.943 & 0.049 & 39.11 & 0.966 & 0.041 \\
3D PatchGAN & 32.95 & 0.886 & 0.070 & 34.47 & 0.906 & 0.063 & 36.45 & 0.923 & 0.059 & 39.03 & 0.952 & 0.044 \\
\rowcolor{gray!15}
\textbf{Real-ESRGAN} & 33.06 & 0.904 & 0.054 & 34.74 & 0.922 & 0.049 & 36.76 & 0.942 & 0.042 & 39.05 & 0.966 & 0.034 \\
\bottomrule
\end{tabularx}
\end{table*}

\subsubsection{Ablation on Inference Infrastructure}
\label{sec:abl_infra}
\cref{fig:infra} adds the optimizations of \cref{sec:infra} to our full model: A is attention operator fusion, B region wise compilation, and C FFN FP8 with QAT. Ours already uses the frame wise 2D decomposition. \cref{tab:infra_full} in \cref{sec:app_infra} gives the time, memory and quality of each cumulative step at all four resolutions.

\noindent\textbf{Optimized runtime.}
\label{sec:abl_opt}
\cref{fig:infra}(a) reports the end to end speedup, \ie the encoding plus decoding time of Ours for a 22 frame clip divided by that of each runtime. Attention operator fusion gives $1.7$--$1.8\times$ at every resolution by removing the intermediate tensors between normalization, RoPE, window attention and the cross frame merge, and region wise compilation raises it to $2.5\times$ at 720p and 1080p and $2.2\times$ at 2160p. FP8 helps only at high resolution: it raises the speedup to $3.0\times$ at 1080p and $2.7\times$ at 2160p but lowers it to $2.1\times$ at 720p and $1.0\times$ at 480p, because its fixed overhead outweighs the saving at low token counts. FP8 also raises the end to end peak memory by 1.4 to 1.7\,GB, so we recommend enabling FP8 only at resolutions of 1080p and above.

\noindent\textbf{QAT.}
\label{sec:abl_qat}
\cref{fig:infra}(b) reports the PSNR change relative to Ours. FP8 applied directly to the trained model lowers PSNR by 1.35\,dB at 480p and 3.43\,dB at 2160p. QAT recovers it to within 0.13\,dB of Ours at every resolution, \eg 39.07\,dB against 39.05\,dB at 2160p. QAT training is described in \cref{sec:infra}.

\section{Conclusion}
\label{sec:conclusion}

We presented LiveVAE, a streaming video VAE that accelerates both encoding and decoding while preserving a pretrained latent space. Our key finding is that a small spatial neighborhood and a short causal temporal window suffice for both components. Causal window attention enables streaming with a bounded KV cache, while three stage distillation maintains compatibility with existing DiTs without retraining. Attention operator fusion, region wise compilation and FFN FP8 with QAT further reduce inference latency. On MiniMax-H3, LiveVAE encodes $5.2$--$6.2\times$ and decodes $3.0$--$3.3\times$ faster than the official VAE at matching or higher PSNR, and generalizes from training at 720p and below to inference at up to 2160p.

\begin{figure}[!ht]
\centering
\begin{minipage}[t]{0.495\linewidth}\vspace{0pt}\centering
\includegraphics[width=\linewidth,trim={0 97.4bp 0 0},clip]{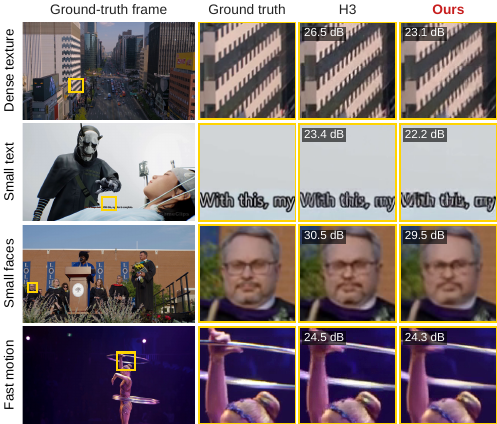}
\end{minipage}\hfill
\begin{minipage}[t]{0.495\linewidth}\vspace{0pt}\centering
\includegraphics[width=\linewidth,trim={0 196.6bp 0 0},clip]{fig/fig_failure.pdf}\\[-0.2pt]
\includegraphics[width=\linewidth,trim={0 0 0 106.35bp},clip]{fig/fig_failure.pdf}
\end{minipage}
\caption{Failure cases. Crops of the ground truth, MiniMax-H3 (H3) and Ours with the PSNR of each crop, on four hard cases: dense texture, small text, small faces and fast motion.}
\label{fig:failure}
\end{figure}

\noindent\textbf{Limitations and future work.}
Despite these gains, LiveVAE retains a perceptual quality gap. Fine details can become blurred (\cref{fig:failure}), and VBench-2.0 Human Fidelity and Physics remain below the official decoder on MiniMax-H3. We attribute part of this gap to a latent space that was learned by a convolutional VAE. Improving perceptual reconstruction and generation quality without increasing streaming latency is a direction for future work.

\bibliographystyle{iclr2026_conference}
\bibliography{references}

\appendix
\crefalias{section}{appendix}

\section{Window Ablation Tables}
\label{sec:app_window}
\cref{tab:window,tab:causal} give the per component numbers behind \cref{fig:window_tradeoff}. Each run replaces one component of MiniMax-H3: the ablated encoder is paired with the official decoder, or the ablated decoder with the official encoder. Time is the per clip p50 latency of the ablated component alone.

\begin{table*}[!ht]
\centering
\footnotesize
\setlength{\aboverulesep}{0pt}\setlength{\belowrulesep}{0pt}\renewcommand{\arraystretch}{1.02}
\setlength{\tabcolsep}{4pt}
\caption{Spatial window size.}
\vspace{-0.8em}
\label{tab:window}
\resizebox{\ifdim\width>\textwidth\textwidth\else\width\fi}{!}{%
\begin{tabular}{l c cccc cccc cccc cccc}
\toprule
\multirow{2}{*}{Module} & \multirow{2}{*}{Window} & \multicolumn{4}{c}{480p} & \multicolumn{4}{c}{720p} & \multicolumn{4}{c}{1080p} & \multicolumn{4}{c}{2160p} \\
\cmidrule(lr){3-6} \cmidrule(lr){7-10} \cmidrule(lr){11-14} \cmidrule(lr){15-18}
 & & PSNR & SSIM & LPIPS & Time & PSNR & SSIM & LPIPS & Time & PSNR & SSIM & LPIPS & Time & PSNR & SSIM & LPIPS & Time \\
\midrule
 & $2\times3\times3$ & 32.67 & 0.899 & 0.079 & 144.0 & 34.23 & 0.917 & 0.071 & 301.1 & 36.08 & 0.937 & 0.060 & 661.4 & 39.49 & 0.966 & 0.045 & 2463.8 \\
\rowcolor{gray!15}
\cellcolor{white} & {\boldmath$2\times7\times7$} & 32.71 & 0.899 & 0.078 & 143.9 & 34.31 & 0.918 & 0.070 & 302.5 & 36.10 & 0.938 & 0.059 & 660.8 & 39.53 & 0.967 & 0.044 & 2464.1 \\
\multirow{-3}{*}{Decoder} & $2\times16\times16$ & 32.84 & 0.901 & 0.077 & 146.8 & 34.45 & 0.919 & 0.069 & 305.5 & 36.26 & 0.939 & 0.058 & 659.6 & 39.67 & 0.967 & 0.043 & 2451.2 \\
\midrule
 & $2\times3\times3$ & 31.57 & 0.883 & 0.065 & 124.4 & 33.10 & 0.903 & 0.058 & 234.1 & 34.87 & 0.926 & 0.050 & 512.7 & 38.19 & 0.958 & 0.038 & 1861.0 \\
\rowcolor{gray!15}
\cellcolor{white} & {\boldmath$2\times7\times7$} & 31.90 & 0.889 & 0.061 & 124.2 & 33.44 & 0.908 & 0.054 & 234.2 & 35.23 & 0.929 & 0.047 & 511.2 & 38.55 & 0.960 & 0.036 & 1855.3 \\
\multirow{-3}{*}{Encoder} & $2\times16\times16$ & 32.02 & 0.890 & 0.059 & 130.9 & 33.56 & 0.909 & 0.053 & 232.8 & 35.17 & 0.929 & 0.046 & 510.8 & 38.21 & 0.959 & 0.036 & 1865.0 \\
\bottomrule
\end{tabular}}
\end{table*}

\begin{table*}[!ht]
\centering
\footnotesize
\setlength{\aboverulesep}{0pt}\setlength{\belowrulesep}{0pt}\renewcommand{\arraystretch}{1.02}
\setlength{\tabcolsep}{4pt}
\caption{Temporal window.}
\vspace{-0.8em}
\label{tab:causal}
\resizebox{\ifdim\width>\textwidth\textwidth\else\width\fi}{!}{%
\begin{tabular}{l c cccc cccc cccc cccc}
\toprule
\multirow{2}{*}{Module} & \multirow{2}{*}{Window} & \multicolumn{4}{c}{480p} & \multicolumn{4}{c}{720p} & \multicolumn{4}{c}{1080p} & \multicolumn{4}{c}{2160p} \\
\cmidrule(lr){3-6} \cmidrule(lr){7-10} \cmidrule(lr){11-14} \cmidrule(lr){15-18}
& & PSNR & SSIM & LPIPS & Time & PSNR & SSIM & LPIPS & Time & PSNR & SSIM & LPIPS & Time & PSNR & SSIM & LPIPS & Time \\
\midrule
\rowcolor{gray!15}
\cellcolor{white} & {\boldmath$2\times7\times7$} & 32.71 & 0.899 & 0.078 & 143.9 & 34.31 & 0.918 & 0.070 & 302.5 & 36.10 & 0.938 & 0.059 & 660.8 & 39.53 & 0.967 & 0.044 & 2464.1 \\
& $3\times7\times7$ & 32.67 & 0.899 & 0.075 & 161.9 & 34.24 & 0.917 & 0.066 & 337.6 & 36.07 & 0.937 & 0.056 & 740.3 & 39.47 & 0.966 & 0.041 & 2755.2 \\
\multirow{-3}{*}{Decoder} & $4\times7\times7$ & 32.87 & 0.901 & 0.076 & 175.9 & 34.45 & 0.919 & 0.068 & 364.6 & 36.30 & 0.939 & 0.058 & 804.2 & 39.75 & 0.967 & 0.043 & 2982.4 \\
\midrule
\rowcolor{gray!15}
\cellcolor{white} & {\boldmath$2\times7\times7$} & 31.90 & 0.889 & 0.061 & 124.2 & 33.44 & 0.908 & 0.054 & 234.2 & 35.23 & 0.929 & 0.047 & 511.2 & 38.55 & 0.960 & 0.036 & 1855.3 \\
& $3\times7\times7$ & 31.96 & 0.889 & 0.060 & 138.2 & 33.51 & 0.909 & 0.054 & 260.6 & 35.29 & 0.930 & 0.047 & 579.5 & 38.57 & 0.961 & 0.035 & 2095.5 \\
\multirow{-3}{*}{Encoder} & $4\times7\times7$ & 31.95 & 0.889 & 0.062 & 153.1 & 33.49 & 0.908 & 0.055 & 285.7 & 35.24 & 0.929 & 0.047 & 631.0 & 38.45 & 0.960 & 0.036 & 2276.9 \\
\bottomrule
\end{tabular}}
\end{table*}
\FloatBarrier

\section{Cumulative Inference Optimizations}
\label{sec:app_infra}
\cref{tab:infra_full} gives the full results behind \cref{fig:infra}: the encoder time (Enc.), the decoder time (Dec.), the end to end peak memory and the quality of each cumulative step. The four evaluation resolutions also show how time and memory scale with the input size.

\begin{table*}[!ht]
\centering
\footnotesize
\setlength{\aboverulesep}{0pt}\setlength{\belowrulesep}{0pt}\renewcommand{\arraystretch}{1.02}
\setlength{\tabcolsep}{4pt}
\caption{Cumulative inference optimizations.}
\vspace{-0.8em}
\label{tab:infra_full}
\begin{tabularx}{\textwidth}{l YYY YYY YYY YYY}
\toprule
\multirow{2}{*}{Runtime} & \multicolumn{3}{c}{480p} & \multicolumn{3}{c}{720p} & \multicolumn{3}{c}{1080p} & \multicolumn{3}{c}{2160p} \\
\cmidrule(lr){2-4} \cmidrule(lr){5-7} \cmidrule(lr){8-10} \cmidrule(lr){11-13}
& Enc. & Dec. & Mem & Enc. & Dec. & Mem & Enc. & Dec. & Mem & Enc. & Dec. & Mem \\
\midrule
Ours & 119.5 & 146.0 & 4.40 & 234.3 & 303.4 & 5.04 & 514.2 & 660.2 & 6.48 & 1856.0 & 2445.0 & 14.12 \\
Ours + A & 74.1 & 85.3 & 4.40 & 130.3 & 174.7 & 5.04 & 281.1 & 376.8 & 6.48 & 1068.4 & 1494.5 & 14.11 \\
Ours + AB & 78.6 & 78.6 & 4.36 & 92.5 & 126.7 & 4.96 & 200.0 & 279.0 & 6.28 & 780.1 & 1135.1 & 13.34 \\
\rowcolor{gray!15}
Ours + ABC & 131.9 & 127.8 & 5.78 & 127.4 & 130.8 & 6.39 & 165.9 & 225.9 & 7.77 & 656.7 & 912.2 & 15.05 \\
\midrule
& PSNR & SSIM & LPIPS & PSNR & SSIM & LPIPS & PSNR & SSIM & LPIPS & PSNR & SSIM & LPIPS \\
\midrule
Ours & 33.06 & 0.904 & 0.054 & 34.74 & 0.922 & 0.049 & 36.76 & 0.942 & 0.042 & 39.05 & 0.966 & 0.034 \\
Ours + ABC w/o QAT & 31.71 & 0.897 & 0.056 & 33.03 & 0.916 & 0.051 & 34.43 & 0.936 & 0.044 & 35.62 & 0.957 & 0.037 \\
\rowcolor{gray!15}
Ours + ABC & 33.00 & 0.903 & 0.054 & 34.66 & 0.921 & 0.048 & 36.63 & 0.941 & 0.041 & 39.07 & 0.966 & 0.032 \\
\bottomrule
\multicolumn{13}{l}{A: attention operator fusion. B: region wise compilation. C: FFN FP8 with QAT.} \\
\end{tabularx}
\end{table*}
\FloatBarrier

\end{document}